\documentclass[10pt,twocolumn,letterpaper]{article}

\usepackage[pagenumbers]{cvpr} 

\usepackage[dvipsnames]{xcolor}

\definecolor{cvprblue}{rgb}{0.21,0.49,0.74}
\definecolor{rosegold}{rgb}{0.90, 0.70, 0.60}
\usepackage[pagebackref,breaklinks,colorlinks,citecolor=rosegold]{hyperref}
\usepackage{multirow}
\usepackage{hhline}
\usepackage{graphicx}
\usepackage{enumitem,kantlipsum}

\def\paperID{24} 
\def\confName{CVPR}
\def\confYear{2024}

\title{Reciprocal Attention Mixing Transformer for Lightweight Image Restoration}

\author{
Haram Choi$^1$\thanks{This work has been done during Master's course in Sogang University.}~~~
Cheolwoong Na$^2$~~~
Jihyeon Oh$^2$~~~
Seungjae Lee$^2$~~~
Jinseop Kim$^2$~~~
Subeen Choe$^2$ \\
Jeongmin Lee$^3$~~~
Taehoon Kim$^4$~~~
Jihoon Yang$^2$\thanks{Corresponding author.}\\
$^1$RippleAI~~~
$^2$Machine Learning Research Lab., Sogang University~~~
$^3$LG Innotek~~~
$^4$LG AI Research
}

\begin{document}
\maketitle
\begin{abstract}
Although many recent works have made advancements in the image restoration (IR) field, they often suffer from an excessive number of parameters.
Another issue is that most Transformer-based IR methods focus only on either local or global features, leading to limited receptive fields or deficient parameter issues.
To address these problems, we propose a lightweight network, Reciprocal Attention Mixing Transformer (RAMiT). 
It employs our proposed dimensional reciprocal attention mixing Transformer (D-RAMiT) blocks, which compute bi-dimensional self-attentions in parallel with different numbers of multi-heads.
The bi-dimensional attentions help each other to complement their counterpart's drawbacks and are then mixed.
Additionally, we introduce a hierarchical reciprocal attention mixing (H-RAMi) layer that compensating for pixel-level information losses and utilizes semantic information while maintaining an efficient hierarchical structure.
Furthermore, we revisit and modify MobileNet V2 to attach efficient convolutions to our proposed components.
The experimental results demonstrate that RAMiT achieves state-of-the-art performance on multiple lightweight IR tasks, including super-resolution, low-light enhancement, deraining, color denoising, and grayscale denoising.
Codes are available at \href{https://github.com/rami0205/RAMiT}{https://github.com/rami0205/RAMiT}.
\end{abstract}    
\section{Introduction}
\label{intro}

\begin{figure*}[t]
\centering
\includegraphics[width=0.85\linewidth]{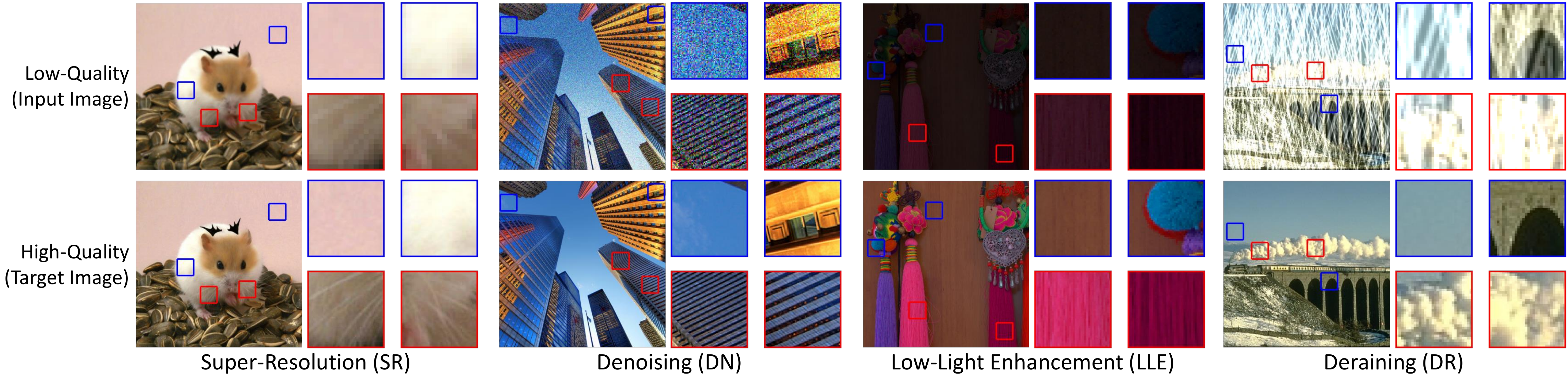}
\caption{The importance of locality and global dependency in image restoration tasks. \textbf{(\textcolor{blue}{Blue} boxes)} Local features are informative enough to recover most parts, meaning that the contribution of locally adjacent pixels is crucial. \textbf{(\textcolor{red}{Red} boxes)} Some areas seem more challenging due to high levels of distortion (blurring, noise, darkness, or obstruction). They require global dependency, which can often be detected in repeated patterns or textures distributed throughout the entire image.}
\label{fig_intro}
\vspace{-10pt}
\end{figure*}

Lightweight image restoration (IR) or enhancement techniques are essential for addressing inherent flaws in images captured in the wild, especially those taken by devices with low computational power.
These techniques aim to reconstruct high-quality images from their distorted low-quality counterparts.
However, many lightweight IR tasks with the popular vision Transformer~\cite{dosovitskiy2021an} based methods remain relatively unexplored.
Although many recent Transformer~\cite{vaswani2017attention} networks have improved the IR domain~\cite{chen2021pre,Zamir2021Restormer,wang2022uformer,zheng2022cross,zhang2023accurate}, they are infeasible for real-world applications due to their large number of parameters.
Furthermore, even the state-of-the-art lightweight IR networks consume intensive computational costs~\cite{liang2021swinir,Lu_2022_CVPR,zhang2022efficient,behjati2023single,choi2023n}.
Another problem is that some IR models mainly focus on expanding the receptive field with respect to locality~\cite{liang2021swinir,wang2022uformer,zhang2022efficient,zheng2022cross,choi2023n}, which is insufficient to capture the global dependency in an image.
This is critical because the IR networks need to refer to repeated patterns and textures distributed throughout the image~\cite{gu2021interpreting,Lu_2022_CVPR}.
Meanwhile, others have tried to enlarge the receptive field globally~\cite{Zamir2021Restormer,zhang2023accurate,behjati2023single} but have overlooked important local (spatial) information, which is conventionally essential for recovery tasks~\cite{he2022masked,wang2022uformer,zheng2022cross,choi2023n}.
Fig.~\ref{fig_intro} visualizes a few examples in which a successful IR depends on the ability to consider both local and global features in a given distorted low-quality image, emphasizing how significant the problem is.

To address these problems, we propose a lightweight IR network called \textbf{RAMiT} (\textbf{R}eciprocal \textbf{A}ttention \textbf{Mi}xing \textbf{T}ransformer).
As shown in Fig.~\ref{fig_overall}\textcolor{red}{a}, RAMiT consists of a shallow module, three hierarchical and the last stages composed of $\mathbb{K}_a$ D-RAMiT blocks before and after an upsampling bottleneck layer, an H-RAMi, and a final reconstruction module.
Our proposed \textbf{D-RAMiT} (\textbf{D}imensional \textbf{R}eciprocal \textbf{A}ttention \textbf{Mi}xing \textbf{T}ransformer) blocks include a novel bi-dimensional self-attention (SA) mixing module.
This operates spatial and channel SA mechanisms~\cite{liang2021swinir,Zamir2021Restormer} in parallel with different multi-heads, and mixes them.
To overcome the drawbacks of each SA, we allow the results from the previous block to help the respective counterparts' SA procedures.
Consequently, RAMiT can capture both local and global dependencies.
Additionally, we propose an efficient component, \textbf{H-RAMi} (\textbf{H}ierarchical \textbf{R}eciprocal \textbf{A}ttention \textbf{Mi}xer) that mixes the multi-scale attentions resulting from four hierarchical stages.
This component complements pixel-level information losses caused by downsampled features, and enhances semantic-level representations.
It enables RAMiT to re-think where and how much attention to pay in the given input feature maps.
For a mixture of each reciprocal (\emph{i.e.,} dimensional and hierarchical) attention result, we modify MobileNet V2~\cite{sandler2018mobilenetv2}.
Utilizing this \textbf{MobiVari} (\textbf{Mobi}leNet \textbf{Vari}ant) layer, we can efficiently and effectively attach the convolutions to the network.

The experimental results demonstrate that the various lightweight IR works are improved by our RAMiT.
As a result, we establish state-of-the-art performance on five different lightweight IR tasks, including super-resolution, low-light enhancement, deraining, color denoising, and grayscale denoising, showing applicability of RAMiT to general low-level vision tasks.
Notably, RAMiT achieves these results with fewer operations or parameters than the other networks.

The summaries of our main contributions are as follows:

\begin{enumerate}[label=(\arabic*), left=0.25cm]
    \item {We propose a dimensional reciprocal attention mixing Transformer (\textbf{D-RAMiT}) block. The spatial and channel self-attentions with the different numbers of multi-heads operate in parallel using and are fused. Therefore, the network can capture both local and global context, which is critical for image restoration tasks.}
    \item {A hierarchical reciprocal attention mixing (\textbf{H-RAMi}) layer is introduced. It compensates for pixel-level information losses caused by downsampled features of hierarchical structure, and utilizes semantic-level information, while maintaining an efficient hierarchical structure.}
    \item {Our RAMiT achieves \textbf{state-of-the-art} results on five different lightweight image restoration tasks. It is noteworthy that RAMiT requires fewer parameters or operations compared to existing methods.}
\end{enumerate}
\section{Related Work}
\label{related}

\textbf{Window Self-Attention.}
After Vision Transformer (ViT)~\cite{dosovitskiy2021an} appeared, Swin Transformer~\cite{liu2021swin} proposed window self-attention (WSA) to solve the excessive time complexity of ViT.
Self-attention is computed with the tokens in a non-overlapping local window.
However, since the receptive field of WSA was limited within a small window, some following high-level vision studies tried to overcome this issue.
GGViT~\cite{yu2021glance}, CrossFormer~\cite{wang2021crossformer}, and MaxViT~\cite{tu2022maxvit} utilized dilated windows to capture the dependency in non-local regions.
Focal Transformer~\cite{yang2021focal} gradually widened surrounding regions (\textit{key, value}) of a local window (\textit{query}).
CSwin\cite{dong2022cswin} extended square windows to cross-shaped rectangle windows.
VSA~\cite{zhang2022vsa} dynamically varied the window size, breaking the local constraint.
DaViT~\cite{ding2022davit} alternately placed spatial WSA and channel self-attention blocks to consider both local and global dependencies in an image.
\vspace{-7pt}

\textbf{WSA for Image Restoration.}
The image restoration (IR) tasks aim to recover a high-quality image from a degraded low-quality counterpart.
SwinIR~\cite{liang2021swinir} firstly adapted window self-attention (WSA) in this domain and achieved outstanding results.
Thereafter, many studies employed WSA and overcame the limited receptive field.
Uformer~\cite{wang2022uformer} proposed locally-enhanced feed-forward network to refer to neighbor pixels.
ELAN~\cite{zhang2022efficient} split channels of input feature maps into different sized windows, efficiently enlarging the local receptive field.
Following \cite{yu2021glance, wang2021crossformer, tu2022maxvit}, ART~\cite{zhang2023accurate} exploited the dilated window attention.
NGswin~\cite{choi2023n} introduced an N-Gram method helping WSA to consider surrounding pattern and texture.
Moreover, Restormer~\cite{Zamir2021Restormer} and NAFNet~\cite{chen2022simple} utilized channel-attention rather than spatial WSA for maximizing the capability of attention mechanism in capturing global dependency.
Related to the approaches above, we aim to address the weakness of the plain WSA.
\section{Methodology}
\label{method}

\begin{figure*}[ht]
\centering
\begin{tabular}{ccc}
    \includegraphics[width=0.4670\linewidth]{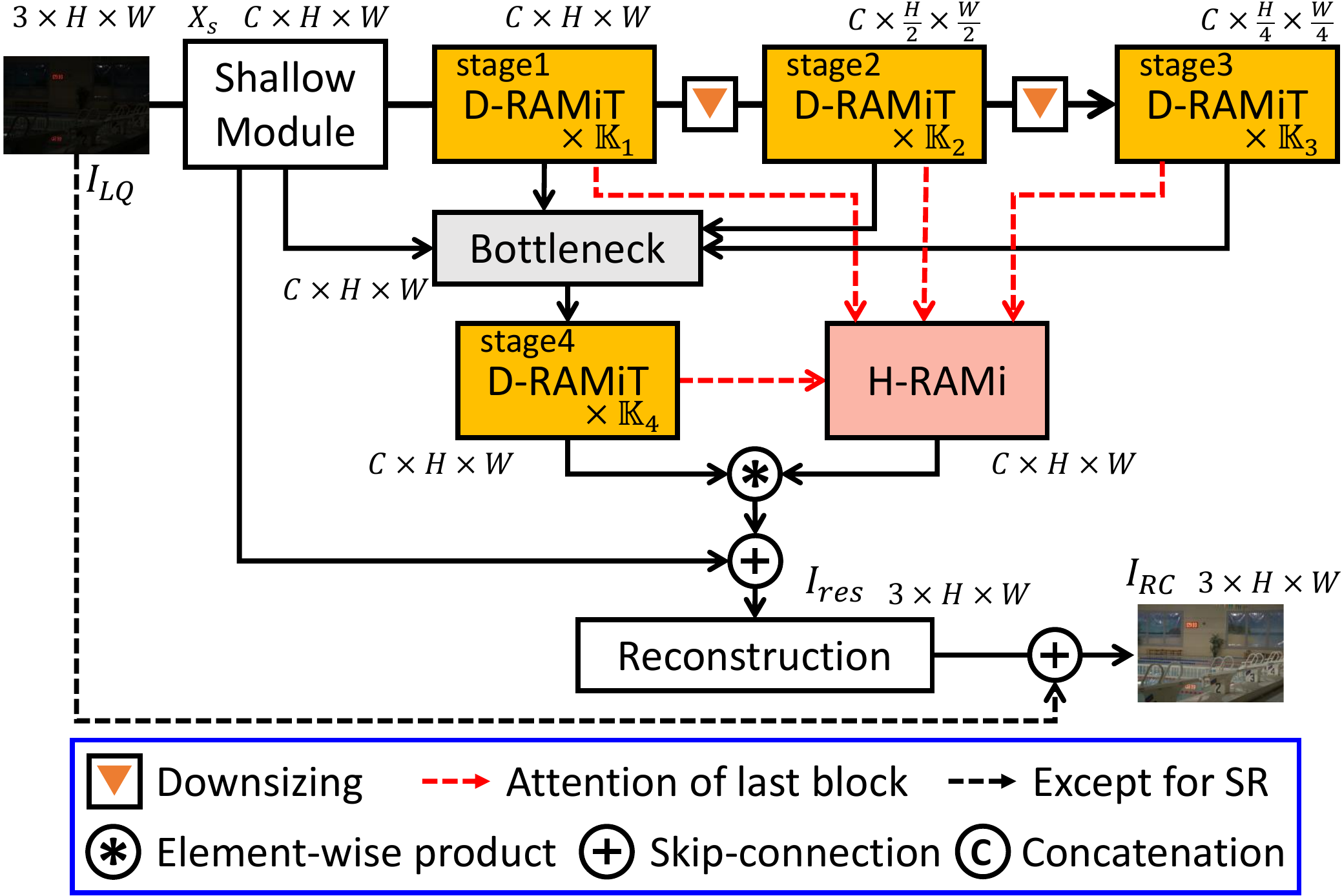} & \hspace{-10pt} \includegraphics[width=0.1741\linewidth]{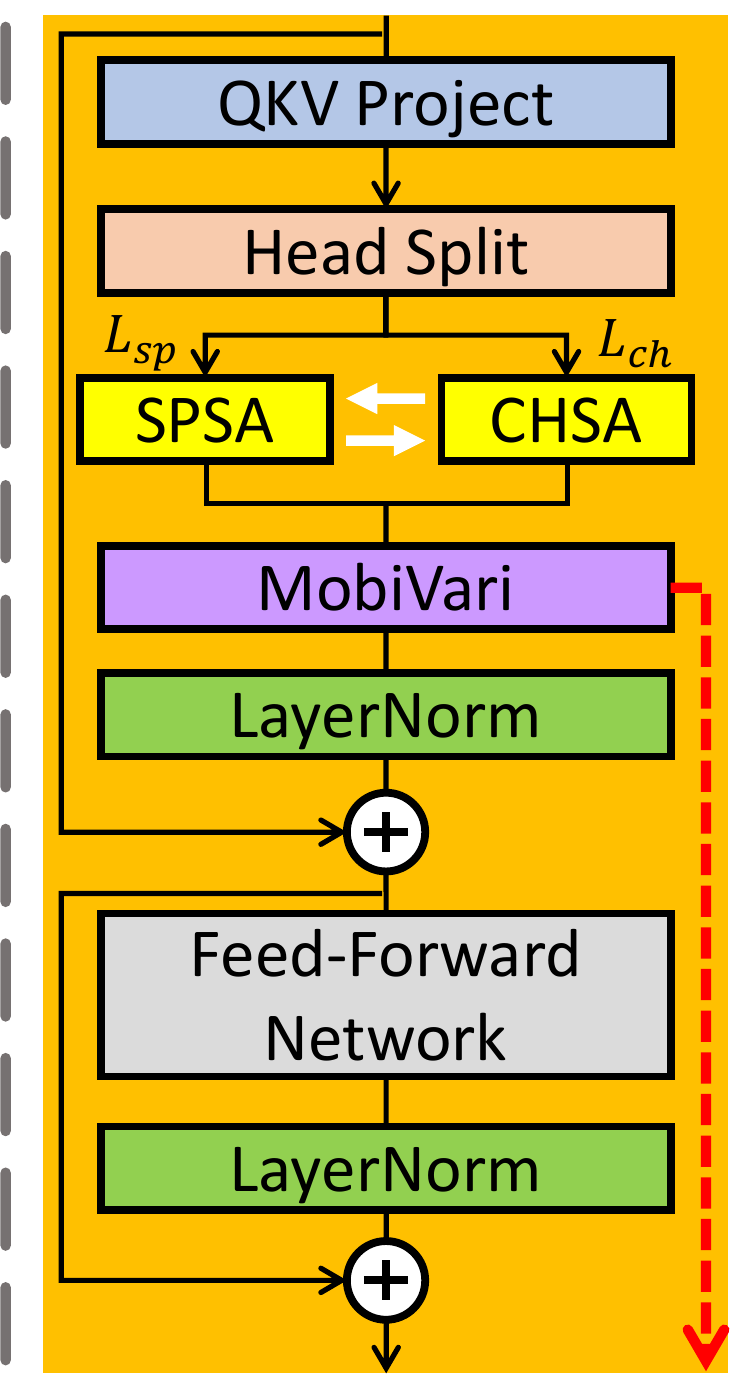} & \hspace{-17pt} \begin{tabular}[b]{c} \includegraphics[width=0.1146\linewidth]{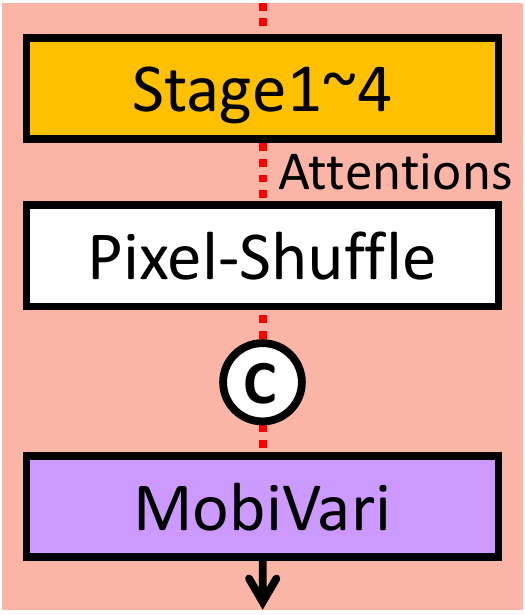} \vspace{-2pt} \\ {\scriptsize (c) H-RAMi} \vspace{8pt} \\ \includegraphics[width=0.1146\linewidth]{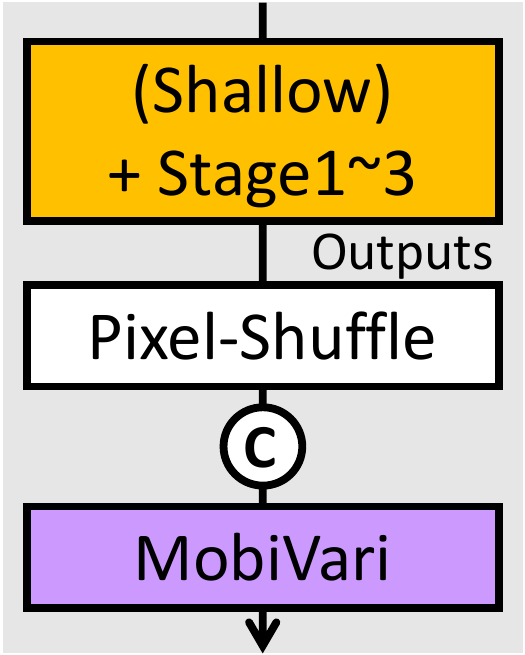} \\ \end{tabular} \hspace{-10pt} \\
    {\scriptsize (a) Overall Architecture of RAMiT} & \hspace{-10pt} {\scriptsize (b) D-RAMiT Block} & \hspace{-17pt} {\scriptsize (d) Bottleneck} \\
\end{tabular}
\caption{Overall architecture of RAMiT. \textbf{(a)} The size indicates dimension of output from each component. The operation of $I_{LQ}+I_{res}$ is omitted for super-resolution tasks. $I_{RC}$ equals to $I_{res}\in\mathbb{R}^{3\times rH \times rW}$ ($r$: an upscale factor). \textbf{(b)} The different multi-heads ($L_{sp}, L_{ch}$) are assigned to each self-attention (SA) module. Being multiplied to \textit{value} of each counterpart, both SAs help each other (white arrows, optional depending on tasks). The bi-dimensional attentions are mixed by our MobileNet variant, MobiVari$^{\textcolor{red}{\text{1}}}$. \textbf{(c)} H-RAMi mixes the hierarchical attentions resulting from the last blocks of each stage. Before MobiVari enhances and mixes the attentions, this module upsamples and concatenates multi-scale attentions. \textbf{(d)} Our bottleneck adopts the SCDP bottleneck of NGswin~\cite{choi2023n}.} 
\vspace{-12pt}
\label{fig_overall}
\end{figure*}

\subsection{Overall Architecture of RAMiT}
\label{overall}

As shown in Fig.~\ref{fig_overall}\textcolor{red}{a}, given a low-quality image $I_{LQ} \in \mathbb{R}^{3 (\text{or} 1) \times H \times W}$, a $3 \times 3$ convolutional \textbf{shallow module} produces $X_s \in \mathbb{R}^{C \times H \times W}$, where $H$ and $W$ are height and width of $I_{LQ}$, and $C$ is channel.
$X_s$ passes through hierarchical \textbf{encoder stages} consisting of $\mathbb{K}_a$ \textbf{D-RAMiT} (\textbf{D}imensional \textbf{R}eciprocal \textbf{A}ttention \textbf{Mi}xing \textbf{T}ransformer, Sec.~\ref{drami} and Fig.~\ref{fig_overall}\textcolor{red}{b}) blocks, where $a$ indicates the stage number.
D-RAMiT calculates self-attention (SA) in bi-dimensions (spatiality and channel) with the different numbers of multi-heads.
After projecting \textit{query, key, value} and splitting current feature map into $L$ heads, $L_{sp}$ and $L_{ch} (=L-L_{sp})$ heads are assigned to spatial and channel self-attention modules, respectively.
For both SAs, we employ scaled-cosine attention and post-normalization~\cite{liu2022swin}.
The reciprocally computed attentions are mixed by \textbf{MobiVari}\footnote{MobiVari modifies the activation function and residual connections and the expansion convolution of the original MobileNet V2~\cite{sandler2018mobilenetv2}. We detail the MobiVari structure in Appendix Sec.~\ref{appendix_mobivari_reconstruction}.} (\textbf{Mobi}leNet \textbf{Vari}ants).
Afterwards, the output passes through layer-norm (LN)~\cite{ba2016layer} with skip connection~\cite{he2016deep}, feed-forward network, and LN.
At the end of the first and second stages, we downsample the feature maps by half, but maintain the channels.
While the \textbf{downsizing layers} follow the patch-merging practice of Swin Transformers~\cite{liu2021swin}, we replace a plain linear projection of these layers with our MobiVari.

When the stage3 ends, $X_s$ and multi-scale outputs from stage $1,2,3$ are fed into a \textbf{bottleneck layer} (Fig.~\ref{fig_overall}\textcolor{red}{d}), which is the same as SCDP bottleneck from NGswin~\cite{choi2023n} except that depth- and point-wise convolution switches over to our MobiVari.
The bottleneck taking multi-scale features can compensate for information loss caused by the downsizing layers.
Using a bottleneck output, the \textbf{stage4} composed of $\mathbb{K}_4$ D-RAMiT blocks operates in the same way as the other stages.
Then, the merged attention results outputted by the last Transformer blocks of all the stages are conveyed to an \textbf{H-RAMi} layer (\textbf{H}ierarchical \textbf{R}eciprocal \textbf{A}ttention \textbf{Mi}xer, Sec.~\ref{hrami} and Fig.~\ref{fig_overall}\textcolor{red}{c}).
H-RAMi upsamples them into $H \times W$ using a pixel-shuffler~\cite{shi2016real} and aggregates them, which is merged by MobiVari.
This layer is simple but robust to pixel-level information losses as is our bottleneck.
The re-mixed hierarchical attention is element-wise multiplied to the stage4 output.
A global skip-connection adds the result with $X_s$~\cite{kim2016accurate}, which is then fed into the \textbf{reconstruction module} to produce a residual image $I_{res}$.
The reconstruction module follows the common practice~\cite{ahn2018fast,liang2021swinir,choi2023n}, but places two MobiVari layers before the original version to boost the performances (detailed in Appendix Sec.~\ref{appendix_mobivari_reconstruction}).
Finally, $I_{res}+I_{LQ}$ makes a reconstructed image $I_{RC}$ (ignored for super-resolution, \emph{i.e.}, $I_{res}=I_{RC}$).

\begin{figure*}[t]
\centering
\begin{subfigure}[h]{0.3\linewidth}
    \includegraphics[width=\linewidth]{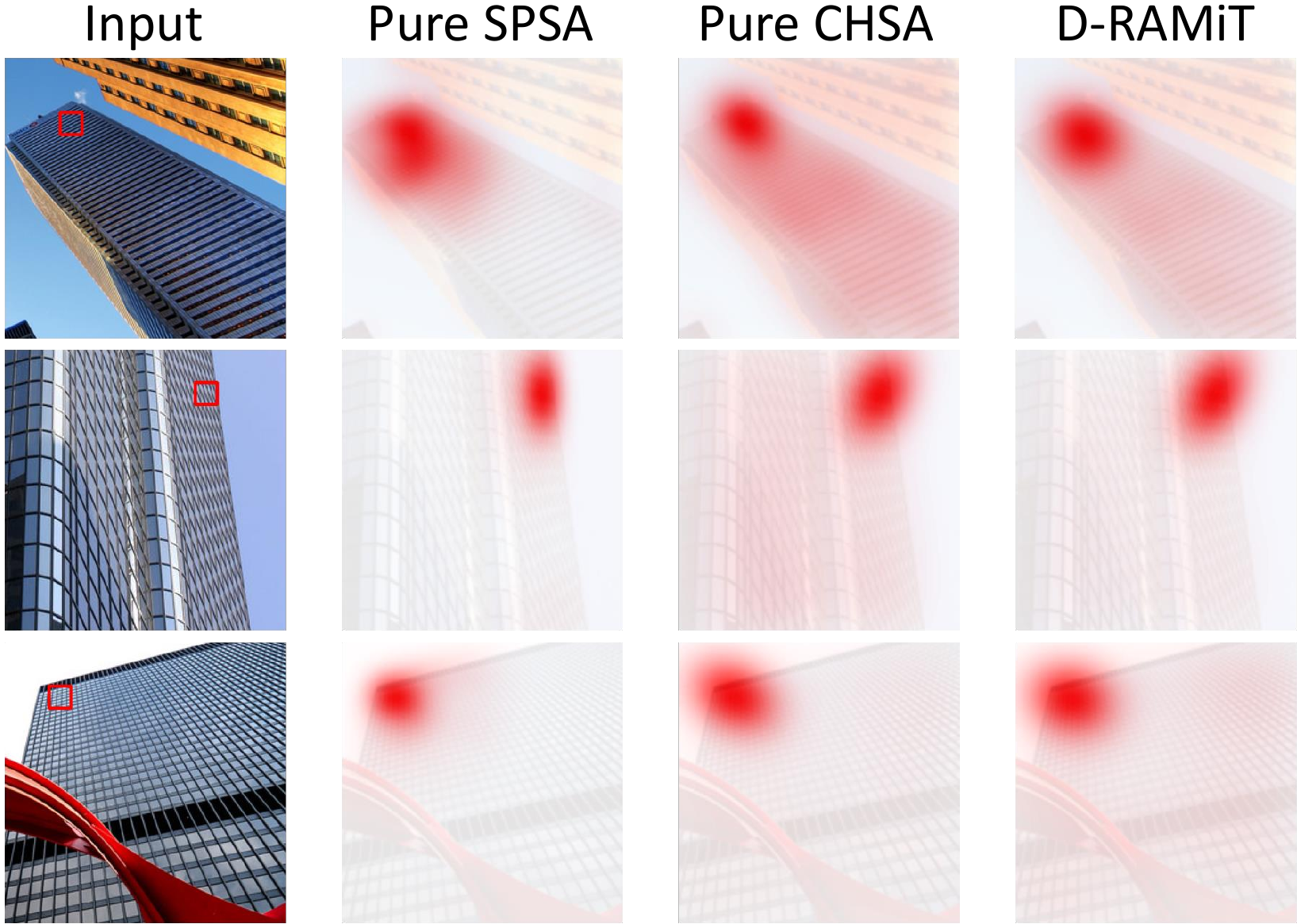}
    \caption{Local Attribution MAP (LAM)~\cite{gu2021interpreting}.}
    \label{fig_lam}
\end{subfigure}
\begin{subfigure}[h]{0.5\linewidth}
    \includegraphics[width=\linewidth]{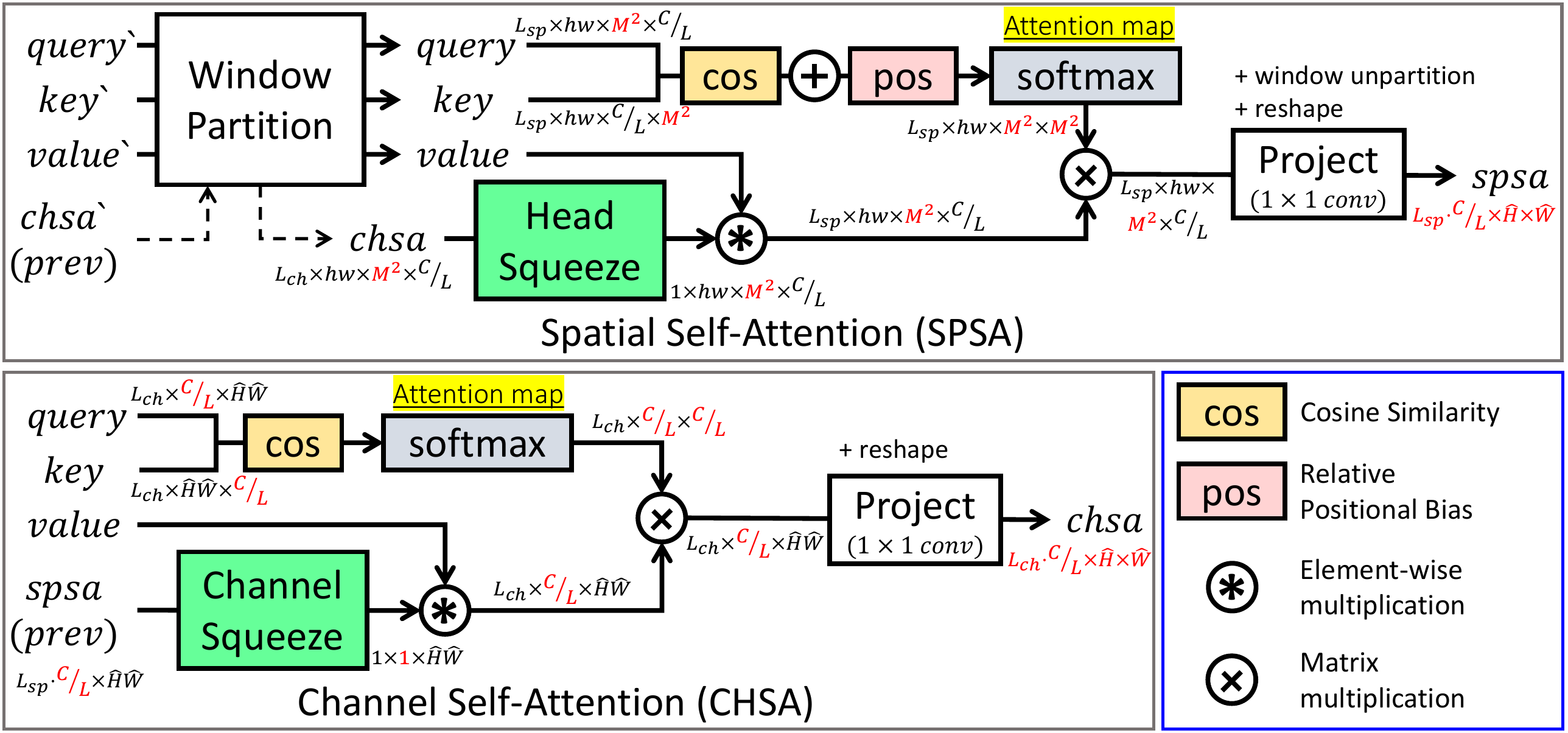}
    \caption{Pipelines of our dimensional reciprocal self-attentions.}
    \label{fig_sa}
\end{subfigure}
\caption{\textbf{(a)} The depth of the red areas indicates the extent to which the regions contribute to recovering a \textcolor{red}{red} box of an input. D-RAMiT utilizes both local and global dependencies, meaningfully expanding the receptive field compared to the pure SPSA (see Appendix Sec.~\ref{appendix_lam}). \textbf{(b)} Our bi-dimensional self-attention schemes help each other to further boost image restoration performances.}
\label{fig_lamsa}
\vspace{-5pt}
\end{figure*}

\subsection{Dimensional Reciprocal Attention Mixing Transformer Block}
\label{drami}

\textbf{Motivation.}
To improve low-level vision tasks like image restoration (IR), it is crucial to refer to repeated patterns and textures distributed through an entire image (\emph{i.e.}, global or non-local context)~\cite{gu2021interpreting,Lu_2022_CVPR}, as already presented in Fig.~\ref{fig_intro}.
Nevertheless, while many approaches for high-level vision tasks, such as classification, have enriched non-locality~\cite{yu2021glance,wang2021crossformer,tu2022maxvit,ding2022davit}, most lightweight IR methods lack the capability to capture global dependency.
They maximize only ``locality'' by adding correlation of adjacent neighbors to a local window~\cite{choi2023n}, or splitting the channels into three groups and the corresponding sizes of local windows within which the self-attention is computed~\cite{zhang2022efficient}.
Meanwhile, channel-attention mechanism is theoretically capable of equipping global dependency by involving all pixels along the channel dimension~\cite{hu2018squeeze,zhang2018image,Zamir2021Restormer,chen2022simple}.
Fig.~\ref{fig_lam} visualizes the actual receptive field of different self-attention methods using the Local Attribution Map~\cite{gu2021interpreting}.
The channel self-attention (\textbf{\underline{CHSA}}) views nearly global areas but performs poorly (Tab~\ref{tab_ablation0}), because lightweight CHSA focuses on the unnecessary parts with deficient trainable parameters~\cite{choi2023exploration}.
On the other hand, the spatial self-attention (\textbf{\underline{SPSA}}\footnote{In this paper, SPSA indicates the local window-based self-attention proposed by Swin Transformer~\cite{liu2021swin}.}) suffers from the limited receptive field despite intensive computational costs (Tab~\ref{tab_ablation0}), which suggests the potential for further improvement.
Hence, our goal is to incorporate local and global context rather than merely enlarging ``local'' receptive field.

\textbf{Proposed Method.}
We propose a bi-dimensional reciprocal self-attention, which is implemented by operating both SPSA and CHSA in parallel (Fig.~\ref{fig_overall}\textcolor{red}{b}).
Our proposed method can capture both local and global range dependency, thereby improving the IR performances.
As illustrated in Fig.~\ref{fig_sa}, our SPSA and CHSA pipelines adapt the local window self-attention of SwinIR~\cite{liang2021swinir} and the transposed attention of Restormer~\cite{Zamir2021Restormer}, respectively.
We assign the different numbers of multi-heads $L_{sp}$ and $L_{ch}$ ($L_{sp}+L_{ch}=L$) to SPSA and CHSA to compute reciprocal attention \textit{Attn}, as follows:
\begin{equation}
\label{eq_dramit}
\begin{aligned}
    & \text{\textit{Attn}} = \text{MobiVari}(\texttt{Concat}[\text{\textit{SPSA}},\text{\textit{CHSA}}])
\end{aligned}
\end{equation}
Each self-attention and the corresponding heads are obtained by Eq.~\ref{eq_attn} and Eq.~\ref{eq_head}, respectively:
\begin{equation}
\label{eq_attn}
\begin{aligned}
    & \text{\textit{SPSA}} = \mathcal{P}_{sp}(\texttt{Concat}[head_1^{sp}, ..., head_{L_{sp}}^{sp}]), \, \\
    & \text{\textit{CHSA}} = \mathcal{P}_{ch}(\texttt{Concat}[head_{L_{sp}+1}^{ch}, ..., head_L^{ch}]) \\ 
\end{aligned}
\end{equation}
\begin{equation}
\label{eq_head}
\begin{aligned}
    & head_i^{sp} = \text{\textit{Softmax}}(cos(Q_i^{sp}, (K_i^{sp})^T)/\tau + B)V_i^{sp}, \\
    & head_i^{ch} = \text{\textit{Softmax}}(cos(Q_i^{ch}, (K_i^{ch})^T)/\tau)V_i^{ch}
\end{aligned}
\end{equation}
$Q_i^{sp}, K_i^{sp}, V_i^{sp}$ and $Q_i^{ch}, K_i^{ch}, V_i^{ch}$ are \textit{query}, \textit{key}, \textit{value} for SPSA and CHSA, respectively;
$cos$ calculates cosine similarity~\cite{liu2022swin};
$B \in \mathbb{R}^{M^2 \times M^2}$ is the relative positional bias~\cite{liu2021swin};
$\tau$ is a trainable scalar that is set larger than 0.01~\cite{liu2022swin};
$\mathcal{P}_{sp}$, $\mathcal{P}_{ch}$ denotes the reshape and projection layer.
Similar to our work, DaViT~\cite{ding2022davit} has sequentially placed the same numbers of SPSA and CHSA blocks.
However, it can consider global context only after attending to spatial dimension (see Appendix Sec.~\ref{appendix_davit}).
In contrast, D-RAMiT processes both SAs in parallel, allocating more heads to SPSA (\emph{e.g.}, $L_{sp}$$:$$L_{ch}$$=$$75\%$$:$$25\%$).
Then, our MobiVari mixes local and global attentions as well as enhances locality by $3\times3$ depth-wise convolution~\cite{Zamir2021Restormer,wang2022uformer}.
The subsequent process follows Fig.~\ref{fig_overall}\textcolor{red}{b}.

\textbf{Reciprocal Helper.}
Our bi-dimensional modules help each other to compensate for each others' weaknesses, thereby further boosting lightweight IR performances.
When operating SPSA of $\ell\text{-}th$ block, \textit{value} is element-wise multiplied with the CHSA output of $(\ell-1)\text{-}th$ block, before multiplying attention map\footnote{Following~\cite{zhang2022efficient}, we remove the attention mask to avoid inefficiency when a cyclic shift~\cite{liu2021swin} is operated.} and \textit{value}.
The inverse process applies to CHSA as well.
It is noteworthy that intensities of information on each SA differ.
Each single channel from the previous CHSA has various global representations.
Thus, we squeeze (average-pool) it at head dimension before product.
On the other hand, averaging channels of SPSA can preserve valuable local properties.
As a result, we squeeze feature of the previous SPSA at channel dimension.
The first D-RAMiT block of each stage excludes this step due to absence of the previous features with a same resolution.
We verify the effects of this approach in Tab.~\ref{tab_ablation_helper}.

\begin{table}
    \centering
    \begin{subtable}[h]{0.5\linewidth}
        \centering
        \resizebox{\linewidth}{!}
        {
        \begin{tabular}{l}
            {\scriptsize $\Omega(\text{\textit{SPSA}}) = 4\hat{H}\hat{W}C^2 + 2M^2\hat{H}\hat{W}C$} \\
            {\scriptsize $\Omega(\text{\textit{CHSA}}) = 4\hat{H}\hat{W}C^2 + 2\hat{H}\hat{W}C^2/L$ }\\
        \end{tabular}
        }
    \end{subtable}
    \begin{subtable}[h]{0.8\linewidth}
        \centering
        \resizebox{\linewidth}{!}
        {
        \begin{tabular}{rrrr}
            \hline
            Task & Pure CHSA & Pure SPSA & \textbf{D-RAMiT} (proposed) \\
            \hline
            SR $\times2$ & 153.4G / 957K & 173.4G / 975K & \textbf{163.4G} / \textbf{940K} \\
            SR $\times4$ & 39.6G / 978K & 44.6G / 996K & \textbf{42.1G} / \textbf{961K} \\
            Denoising & 583.2G / 952K & 659.9G / 970K & \textbf{620.8G} / \textbf{935K} \\
            \hline
            \multicolumn{3}{l}{{\small *SR: Super-Resolution}} \\
            \multicolumn{3}{l}{{\small *Both methods have the same number of layers and channels.}} \\
        \end{tabular}
        }
    \end{subtable}
    \caption{\textbf{(Eq.)} Time complexity. \textbf{(Tab.)} Mult-Adds / \#Parameters.}
    \label{tab_complexity}
    \vspace{-10pt}
\end{table}

\begin{figure*}[t]
\centering
\includegraphics[width=0.75\linewidth]{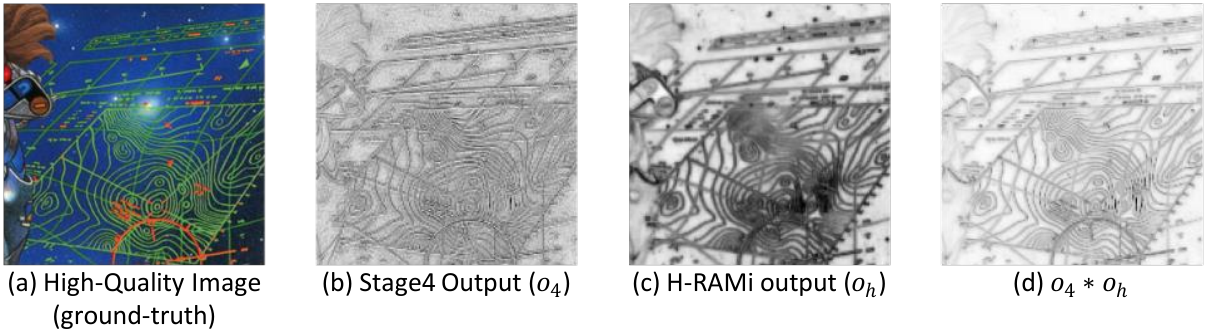}
\caption{Impacts of H-RAMi. \textbf{(a)} A ground-truth high-quality image. \textbf{(b)}, \textbf{(c)} The feature maps after stage $4$ and H-RAMi. \textbf{(d)} Element-wise product of (b) and (c) (Remind Fig.~\ref{fig_overall}\textcolor{red}{a}). (b), (c), (d) are obtained by max-pooling along channel and standardization. More are in Appendix Sec.~\ref{appendix_hrami}.}
\label{fig_hrami}
\vspace{-10pt}
\end{figure*}

\textbf{Efficiency.}
The pure SPSA module employed by other IR networks~\cite{liang2021swinir,zhang2022efficient,choi2023n} have quadratic time complexity to a local window size.
On the other hand, the time complexity of a CHSA module is usually lower than that of an SPSA, as channels per head ($C/L$) is mostly not larger than a local window area ($M^2$) in the equations of Tab.~\ref{tab_complexity}.
Our proposed D-RAMiT, thus, is more efficient than the pure SPSA.
Moreover, D-RAMiT significantly compensates the limited capability of the pure CHSA (see Tab.~\ref{tab_ablation0}).
Mult-Adds is evaluated on a $1280\times720$ high-resolution image.

\subsection{Hierarchical Reciprocal Attention Mixer}
\label{hrami}

\textbf{Motivation.}
There are many evidences that a hierarchical network is usually less effective for IR tasks~\cite{choi2023n,choi2023exploration,hou2020strip,zhao2020hierarchical}.
This is because the goal of IR is to predict pixel values one by one (\emph{i.e.}, dense prediction) inferring recovery patterns when given the distribution of other pixels~\cite{gu2021interpreting}.
However, downsizing feature maps significantly loses important pixel-level information, which prevents many IR researchers from employing hierarchical structures~\cite{zhang2018image,ahn2018fast,luo2020latticenet,niu2020single,zhang2022efficient,behjati2023single}.
Nevertheless, a hierarchical architecture has several advantages.
First, reducing the feature map size can lower time complexity.
For example, non-hierarchical SwinIR-light~\cite{liang2021swinir} requires intensive computations (See Tab.~\ref{tab_sr}).
Furthermore, a hierarchical structure can learn semantic-level feature representation as well as pixel-level~\cite{hariharan2015hypercolumns,wang2022restoring}.
To complement the demerits and leverage the merits, we propose the Hierarchical Reciprocal Attention Mixing layer.

\textbf{Proposed Method.}
As presented in Fig.~\ref{fig_overall}\textcolor{red}{c}, our H-RAMi layer is simple but effective.
Inspired by SCDP bottleneck~\cite{choi2023n}, we apply the same strategy to ``multi-scale attentions'' from the hierarchical encoder stages instead of the final outputs.
H-RAMi takes the attentions merged by MobiVari before layer-norm~\cite{ba2016layer} (a \textcolor{red}{red} dashed arrow next to a \textcolor{violet}{violet} rectangle of Fig.~\ref{fig_overall}\textcolor{red}{b}) of the last D-RAMiT blocks in the hierarchical stage $1,2,3,4$.
After we upsample the resolutions of the mixed bi-dimensional attentions (inputs) into $H \times W$, they are concatenated and mixed by our MobiVari.
Therefore, our H-RAMi can take advantage of both multi-scale and bi-dimensional attentions, re-considering where and how much attention to pay semantically and globally.
Fig.~\ref{fig_hrami} illustrates the impacts of H-RAMi.
The output of stage $4$ at (b) produces relatively unclear edges for fine-grained areas.
This vulnerability stems from less abundant pixel-level information than non-hierarchical networks~\cite{liang2021swinir,zhang2022efficient,behjati2023single}.
However, H-RAMi reconstructs attentive areas and produces clearer borders at (c) by taking both pixel- and semantic-level information.
As a result, the re-attended feature map at (d) contains more apparent and obvious boundaries, which enhances the image restoration accuracy (Tab.~\ref{tab_ablation0}).
\section{Experiments}
\label{experiments}

\subsection{Experimental Setup}
\label{setting}

{\bf Training.}
We randomly cropped low-quality (LQ) images into various sizes of patches according to each task.
The training data was augmented by the random horizontal flip and rotation ($90^{\circ}$, $180^{\circ}$, $270^{\circ}$) as done in the recent works~\cite{liang2021swinir,zhang2022efficient,behjati2023single,choi2023n}.
We minimized $L_1$ pixel-loss between $I_{RC}$ and a ground truth high-quality image $I_{HQ}$: $\mathcal{L} = \lVert I_{HQ}-I_{RC} \rVert_1$ with Adam~\cite{kingma2014adam} optimizer.
For image super-resolution (\underline{SR}), 800 high and low resolution image pairs from DIV2K~\cite{agustsson2017ntire} dataset were used.
The low-resolution images were acquired by the MATLAB bicubic kernel from corresponding high-resolution images.
The color and grayscale image denoising (\underline{DN}) models were trained on DFBW, a merged dataset of 800 DIV2K, 2,650 Flickr2K~\cite{timofte2017ntire}, 400 BSD500~\cite{arbelaez2010contour}, and 4,744 WED~\cite{ma2016waterloo} images, following~\cite{liang2021swinir,Zamir2021Restormer,zhang2023accurate,choi2023exploration}.
The random Gaussian noise level $\sigma$ ranging $[0,50]$ was used to get noisy LQ images.
For low-light image enhancement (\underline{LLE}), 1,785 dark and bright image pairs were utilized (485 LOL~\cite{Chen2018Retinex} + 1,300 VE-LOL~\cite{liu2021benchmarking}), which were either captured or synthesized.
Next, we trained our deraining (\underline{DR}) model on 13,711 synthesized rainy and clean image pairs of Rain13K~\cite{Zamir2021MPRNet} collected from~\cite{fu2017removing, yang2017deep, zhang2019image, zhang2018density, li2016rain}.
Other details are in Appendix Sec.~\ref{appendix_setting}.

\begin{figure*}[t]
    \centering
    \includegraphics[width=0.7441\linewidth]{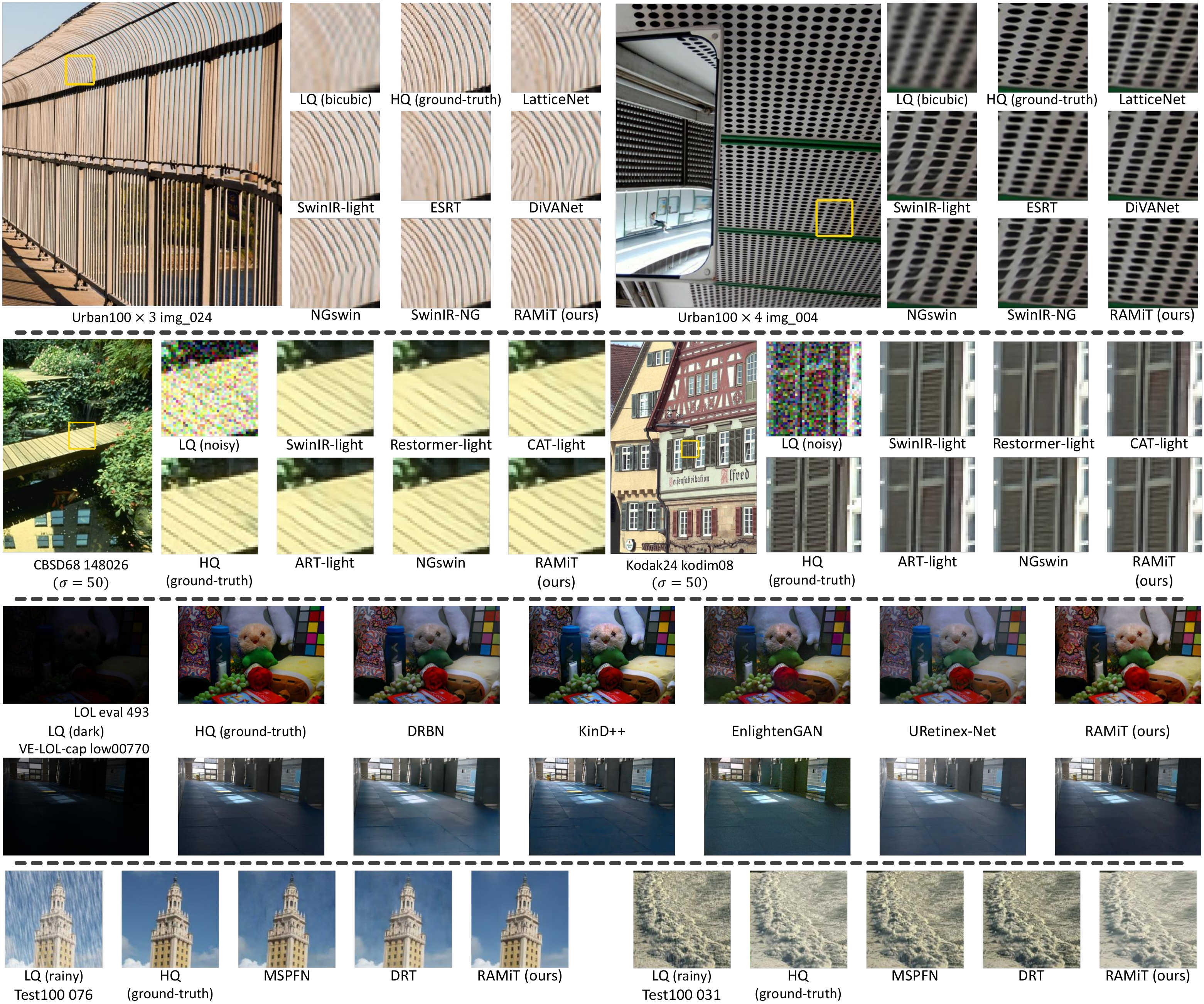} \\
    \caption{Visual comparisons of multiple lightweight image restoration tasks. LQ: Low-Quality input. HQ: High-Quality target. \textbf{(1st row)} Super-Resolution. \textbf{(2nd row)} Denoising. \textbf{(3rd row)} Low-Light Enhancement. \textbf{(4th row)} Deraining. More results are provided in Appendix Sec.~\ref{appendix_viscomp}.}
    \label{fig_viscomp}
    \vspace{-10pt}
\end{figure*}

{\bf Evaluation.}
For SR, we evaluated the performances on the five benchmark datasets, composed of Set5~\cite{bevilacqua2012low}, Set14~\cite{zeyde2010single}, BSD100~\cite{martin2001database}, Urban100~\cite{huang2015single}, and Manga109~\cite{matsui2017sketch}.
We calculated PSNR (dB) and SSIM~\cite{wang2004image} scores on the Y channel of the YCbCr space.
The same metrics were calculated for testing DR, which involves Test100~\cite{zhang2019image} and Rain100H~\cite{yang2017deep} datasets.
To test DN performances, Gaussian noise with different levels $\sigma$ of $\{15, 25, 50\}$ is added.
We reported PSNR and SSIM on the RGB channel of CBSD68~\cite{martin2001database}, Kodak24~\cite{franzen1999kodak}, McMaster~\cite{zhang2011color}, and Urban100 for color DN and on Y channel of Set12~\cite{zhang2017beyond}, BSD68~\cite{martin2001database}, and Urban100 for grayscale DN.
The same metrics for color DN were employed to evaluate the LLE performances on 15 LOL~\cite{Chen2018Retinex} and 100 VE-LOL-cap~\cite{liu2021benchmarking} test images.

\subsection{Qualitative Comparisons}
\label{qual}
Fig.~\ref{fig_viscomp} presents the visual comparisons with other models, which were selected based on existing state-of-the-art studies for each task.
The illustration demonstrates that our proposed dimensional and hierarchical attention mixing methods were able to recover more accurate textures and patterns than other methods.
Our combination of ``local and global'' and ``pixel- and semantic-level'' features made our proposed approach effective.
More results are in Appendix Sec.~\ref{appendix_viscomp}.

\subsection{Quantitative Comparisons}
\label{quan}

\begin{table*}[t]
    \centering
    \resizebox{0.75\linewidth}{!}
    {
    \begin{tabular}{|r|cc|c|c|c|c|c|c|c|c|c|c|c|}
        \hline
        \multirow{2}{*}{Method} & \multirow{2}{*}{Mult-Adds} & \multirow{2}{*}{\#Params} & \multirow{2}{*}{Scale} & \multicolumn{2}{c|}{Set5~\cite{bevilacqua2012low}} & \multicolumn{2}{c|}{Set14~\cite{zeyde2010single}} & \multicolumn{2}{c|}{BSD100~\cite{martin2001database}} & \multicolumn{2}{c|}{Urban100~\cite{huang2015single}} & \multicolumn{2}{c|}{Manga109~\cite{matsui2017sketch}} \\ \cline{5-14}
        &&&&PSNR&SSIM&PSNR&SSIM&PSNR&SSIM&PSNR&SSIM&PSNR&SSIM \\
        \hline
        CARN~\cite{ahn2018fast} & 222.8G & 1,592K & \multirow{11}{*}{$\times2$} & 37.76 & 0.9590 & 33.52 & 0.9166 & 32.09 & 0.8978 & 31.92 & 0.9256 & 38.36 & 0.9765 \\
        LatticeNet~\cite{luo2020latticenet} & 169.5G & 756K & & 38.06 & 0.9607 & 33.70 & 0.9187 & 32.20 & 0.8999 & 32.25 & 0.9288 & 38.94 & 0.9774 \\
        SwinIR-light~\cite{liang2021swinir} & 243.7G & 910K & & 38.14 & 0.9611 & 33.86 & 0.9206 & \textcolor{blue}{32.31} & 0.9012 & 32.76 & \textcolor{blue}{0.9340} & 39.12 & \textcolor{red}{0.9783} \\
        FMEN~\cite{du2022fast} & 172.0G & 748K & & 38.10 & 0.9609 & 33.75 & 0.9192 & 32.26 & 0.9007 & 32.41 & 0.9311 & 38.95 & 0.9778 \\
        ESRT~\cite{Lu_2022_CVPR} & 191.4G & \textcolor{blue}{677K} & & 38.03 & 0.9600 & 33.75 & 0.9184 & 32.25 & 0.9001 & 32.58 & 0.9318 & 39.12 & 0.9774 \\
        ELAN-light~\cite{zhang2022efficient} & 168.4G & \textcolor{red}{582K} & & \textcolor{red}{38.17} & 0.9611 & \textcolor{blue}{33.94} & \textcolor{blue}{0.9207} & 32.30 & 0.9012 & 32.76 & \textcolor{blue}{0.9340} & 39.12 & \textcolor{red}{0.9783} \\
        DiVANet~\cite{behjati2023single} & 189.0G & 902K & & \textcolor{blue}{38.16} & \textcolor{red}{0.9612} & 33.80 & 0.9195 & 32.29 & 0.9012 & 32.60 & 0.9325 & 39.08 & 0.9775 \\
        NGswin~\cite{choi2023n} & \textcolor{red}{140.4G} & 998K & & 38.05 & 0.9610 & 33.79 & 0.9199 & 32.27 & 0.9008 & 32.53 & 0.9324 & 38.97 & 0.9777 \\
        SwinIR-NG~\cite{choi2023n} & 274.1G & 1,181K & & \textcolor{red}{38.17} & \textcolor{red}{0.9612} & \textcolor{blue}{33.94} & 0.9205 & \textcolor{blue}{32.31} & \textcolor{blue}{0.9013} & \textcolor{blue}{32.78} & \textcolor{blue}{0.9340} & \textcolor{blue}{39.20} & \textcolor{blue}{0.9781} \\
        \hhline{|=|==|~|=|=|=|=|=|=|=|=|=|=|}
        \textbf{RAMiT (ours)} & \textcolor{blue}{\textbf{163.4G}} & \textbf{940K} & & \textcolor{blue}{\textbf{38.16}} & \textcolor{red}{\textbf{0.9612}} & \textcolor{red}{\textbf{34.00}} & \textcolor{red}{\textbf{0.9213}} & \textcolor{red}{\textbf{32.33}} & \textcolor{red}{\textbf{0.9015}} & \textcolor{red}{\textbf{32.81}} & \textcolor{red}{\textbf{0.9346}} & \textcolor{red}{\textbf{39.32}} & \textcolor{red}{\textbf{0.9783}} \\
        \hline\hline
        CARN~\cite{ahn2018fast} & 118.8G & 1,592K & \multirow{11}{*}{$\times3$} & 34.29 & 0.9255 & 30.29 & 0.8407 & 29.06 & 0.8034 & 28.06 & 0.8493 & 33.50 & 0.9440 \\
        LatticeNet~\cite{luo2020latticenet} & 76.3G & 765K & & 34.40 & 0.9272 & 30.32 & 0.8416 & 29.10 & 0.8049 & 28.19 & 0.8513 & 33.63 & 0.9442 \\
        SwinIR-light~\cite{liang2021swinir} & 109.5G & 918K & & 34.62 & 0.9289 & 30.54 & 0.8463 & 29.20 & 0.8082 & 28.66 & 0.8624 & 33.98 & 0.9478 \\
        FMEN~\cite{du2022fast} & 77.2G & \textcolor{blue}{757K} & & 34.45 & 0.9275 & 30.40 & 0.8435 & 29.17 & 0.8063 & 28.33 & 0.8562 & 33.86 & 0.9462 \\
        ESRT~\cite{Lu_2022_CVPR} & 96.4G & 770K & & 34.42 & 0.9268 & 30.43 & 0.8433 & 29.15 & 0.8063 & 28.46 & 0.8574 & 33.95 & 0.9455 \\
        ELAN-light~\cite{zhang2022efficient} & \textcolor{blue}{75.7G} & \textcolor{red}{590K} & & 34.61 & 0.9288 & 30.55 & 0.8463 & 29.21 & 0.8081 & 28.69 & 0.8624 & 34.00 & 0.9478 \\
        DiVANet~\cite{behjati2023single} & 89.0G & 949K & & 34.60 & 0.9285 & 30.47 & 0.8447 & 29.19 & 0.8073 & 28.58 & 0.8603 & 33.94 & 0.9468 \\
        NGswin~\cite{choi2023n} & \textcolor{red}{66.6G} & 1,007K & & 34.52 & 0.9282 & 30.53 & 0.8456 & 29.19 & 0.8078 & 28.52 & 0.8603 & 33.89 & 0.9470 \\
        SwinIR-NG~\cite{choi2023n} & 114.1G & 1,190K & & \textcolor{red}{34.64} & \textcolor{red}{0.9293} & \textcolor{blue}{30.58} & \textcolor{red}{0.8471} & \textcolor{blue}{29.24} & \textcolor{blue}{0.8090} & \textcolor{blue}{28.75} & \textcolor{blue}{0.8639} & \textcolor{blue}{34.22} & \textcolor{blue}{0.9488} \\
        \hhline{|=|==|~|=|=|=|=|=|=|=|=|=|=|}
        \textbf{RAMiT (ours)} & \textbf{77.3G} & \textbf{949K} & & \textcolor{blue}{\textbf{34.63}} & \textcolor{blue}{\textbf{0.9290}} & \textcolor{red}{\textbf{30.60}} & \textcolor{blue}{\textbf{0.8467}} & \textcolor{red}{\textbf{29.25}} & \textcolor{red}{\textbf{0.8093}} & \textcolor{red}{\textbf{28.76}} & \textcolor{red}{\textbf{0.8646}} & \textcolor{red}{\textbf{34.30}} & \textcolor{red}{\textbf{0.9490}} \\
        \hline\hline
        CARN~\cite{ahn2018fast} & 90.9G & 1,592K & \multirow{11}{*}{$\times4$} & 32.13 & 0.8937 & 28.60 & 0.7806 & 27.58 & 0.7349 & 26.07 & 0.7837 & 30.47 & 0.9084 \\
        LatticeNet~\cite{luo2020latticenet} & 43.6G & 777K & & 32.18 & 0.8943 & 28.61 & 0.7812 & 27.57 & 0.7355 & 26.14 & 0.7844 & 30.54 & 0.9075 \\
        SwinIR-light~\cite{liang2021swinir} & 61.7G & 930K & & \textcolor{blue}{32.44} & 0.8976 & 28.77 & 0.7858 & 27.69 & \textcolor{blue}{0.7406} & 26.47 & 0.7980 & 30.92 & 0.9151 \\
        FMEN~\cite{du2022fast} & 44.2G & 769K & & 32.24 & 0.8955 & 28.70 & 0.7839 & 27.63 & 0.7379 & 26.28 & 0.7908 & 30.70 & 0.9107 \\
        ESRT~\cite{Lu_2022_CVPR} & 67.7G & \textcolor{blue}{751K} & & 32.19 & 0.8947 & 28.69 & 0.7833 & 27.69 & 0.7379 & 26.39 & 0.7962 & 30.75 & 0.9100 \\
        ELAN-light~\cite{zhang2022efficient} & 43.2G & \textcolor{red}{601K} & & 32.43 & 0.8975 & \textcolor{blue}{28.78} & 0.7858 & 27.69 & \textcolor{blue}{0.7406} & 26.54 & 0.7982 & 30.92 & 0.9150 \\
        DiVANet~\cite{behjati2023single} & 57.0G & 939K & & 32.41 & 0.8973 & 28.70 & 0.7844 & 27.65 & 0.7391 & 26.42 & 0.7958 & 30.73 & 0.9119 \\
        NGswin~\cite{choi2023n} & \textcolor{red}{36.4G} & 1,019K & & 32.33 & 0.8963 & \textcolor{blue}{28.78} & 0.7859 & 27.66 & 0.7396 & 26.45 & 0.7963 & 30.80 & 0.9128 \\
        SwinIR-NG~\cite{choi2023n} & 63.0G & 1,201K & & \textcolor{blue}{32.44} & \textcolor{blue}{0.8980} & \textcolor{red}{28.83} & \textcolor{blue}{0.7870} & \textcolor{red}{27.73} & \textcolor{red}{0.7418} & \textcolor{red}{26.61} & \textcolor{blue}{0.8010} & \textcolor{blue}{31.09} & \textcolor{blue}{0.9161} \\
        \hhline{|=|==|~|=|=|=|=|=|=|=|=|=|=|}
        \textbf{RAMiT (ours)} & \textcolor{blue}{\textbf{42.1G}} & \textbf{961K} & & \textcolor{red}{\textbf{32.56}} & \textcolor{red}{\textbf{0.8992}} & \textcolor{red}{\textbf{28.83}} & \textcolor{red}{\textbf{0.7873}} & \textcolor{blue}{\textbf{27.71}} & \textcolor{red}{\textbf{0.7418}} & \textcolor{blue}{\textbf{26.60}} & \textcolor{red}{\textbf{0.8017}} & \textcolor{red}{\textbf{31.17}} & \textcolor{red}{\textbf{0.9170}} \\
        \hline
    \end{tabular}
    }
    \caption{Comparison of lightweight super-resolution results. Mult-Adds is evaluated on a $1280\times720$ high-resolution image. The best and second best results are in \textcolor{red}{red} and \textcolor{blue}{blue}.}
    \label{tab_sr}
    \vspace{-5pt}
\end{table*}

\begin{table*}[t]
    \centering
    \begin{subtable}[h]{0.4094\linewidth}
        \caption{Low-Light Image Enhancement (LLE).}
        \label{tab_lle}
        \centering
        \resizebox{\linewidth}{!}
        {
        \begin{tabular}{|r|c|c|c|c|c|}
        	\hline
            \multirow{2}{*}{Method} & \multirow{2}{*}{\#Params} & \multicolumn{2}{c|}{LOL~\cite{Chen2018Retinex}} & \multicolumn{2}{c|}{VE-LOL-cap~\cite{liu2021benchmarking}} \\
            \cline{3-6}
            & & PSNR & SSIM & PSNR & SSIM \\
            \hline
            DRBN~\cite{yang2020fidelity} & \textcolor{blue}{558K} & 18.80 & 0.8304 & 20.11 & 0.8545 \\
            KinD++~\cite{zhang2021beyond} & 8,275K & \textcolor{blue}{21.80} & 0.8338 & \textcolor{blue}{22.21} & 0.8430 \\
            EnlightenGAN~\cite{jiang2021enlightengan} & 8,640K & 17.48 & 0.6507 & 18.64 & 0.6754 \\
            URetinex-Net~\cite{wu2022uretinex} & \textcolor{red}{361K} & 21.33 & \textcolor{blue}{0.8348} & 21.22 & \textcolor{blue}{0.8593} \\
            \hhline{|=|=|=|=|=|=|}
            \textbf{RAMiT (ours)} & \textbf{935K} & \textcolor{red}{\textbf{24.14}} & \textcolor{red}{\textbf{0.8423}} & \textcolor{red}{\textbf{28.73}} & \textcolor{red}{\textbf{0.8886}} \\
            \hline
        \end{tabular}
        }
    \end{subtable}
    \begin{subtable}[h]{0.3776\linewidth}
        \caption{Image Deraining (DR).}
        \label{tab_dr}
        \centering
        \resizebox{\linewidth}{!}
        {
        \begin{tabular}{|r|c|c|c|c|c|}
            \hline
            \multirow{2}{*}{Method} & \multirow{2}{*}{\#Params} & \multicolumn{2}{c|}{Test100~\cite{zhang2019image}} & \multicolumn{2}{c|}{Rain100H~\cite{yang2017deep}} \\
            \cline{3-6}
            & & PSNR & SSIM & PSNR & SSIM \\
            \hline
            UMRL~\cite{yasarla2019uncertainty} & 984K & 24.41 & 0.8290 & 26.01 & 0.8320 \\
            MSPFN~\cite{jiang2020multi} & 13,350K & 27.50 & 0.8760 & 28.66 & 0.8600 \\
            DRT~\cite{liang2022drt} & 1,180K & 27.02 & 0.8470 & \textcolor{blue}{29.47} & 0.8460 \\
            TAO-Net~\cite{li2022tao} & \textcolor{red}{755K} & \textcolor{blue}{28.59} & \textcolor{blue}{0.8870} & 28.96 & \textcolor{blue}{0.8640} \\
            \hhline{|=|=|=|=|=|=|}
            \textbf{RAMiT (ours)} & \textcolor{blue}{\textbf{935K}} & \textcolor{red}{\textbf{30.44}} & \textcolor{red}{\textbf{0.9012}} & \textcolor{red}{\textbf{29.69}} & \textcolor{red}{\textbf{0.8775}} \\
            \hline
        \end{tabular}
        }
    \end{subtable}
    \caption{Comparison of lightweight low-light image enhancement and image deraining results.}
    \label{tab_drlle}
    \vspace{-5pt}
\end{table*}

\begin{table*}[t]
    \centering
    \begin{subtable}[h]{\linewidth}
        \caption{D-RAMiT \& H-RAMi (Mult-Adds / \#Params / Average PSNR).}
        \label{tab_ablation0}
        \centering
        \resizebox{0.75\linewidth}{!}
        {
        \begin{tabular}{|cc|ccccc|l}
        	\cline{1-7}
            Transformer & H-RAMi & SR $\times2$ & SR $\times4$ & CDN $\sigma=50$ & LLE & DR & \\
            \cline{1-7}
            Pure CHSA & \textit{w/} & 153.4G / 957K / 34.994 & 39.6G / 978K / 29.074 & 583.2G / 952K / 28.848 & 583.2G / 952K / 23.985 & 583.2G / 952K / 28.810 & \{i\} \\
            \cline{1-7}
            Pure SPSA & \textit{w/o} & 168.6G / 955K / 35.218 & 43.4G / 976K / 29.302 & 641.2G / 950K / 29.010 & 641.2G / 950K / 25.095 & 641.2G / 950K / 29.175 & \{ii\} \\
            Pure SPSA & \textit{w/} & 173.4G / 975K / 35.276 & 44.6G / 996K / 29.342 & 659.9G / 970K / 29.128 & 659.9G / 970K / 25.140 & 659.9G / 970K / 29.190 & \{iii\} \\
            \cline{1-7}
            D-RAMiT & \textit{w/o} & 158.5G / 920K / 35.310 & 40.9G / 940K / 29.338 & 602.1G / 914K / 29.205 & 602.9G / 914K / 26.365 & 602.1G / 914K / 29.940 & \{iv\} \\
            D-RAMiT & \textit{w/} & 163.4G / 940K / \textbf{35.324} & 42.1G / 961K / \textbf{29.374} & 620.8G / 935K / \textbf{29.275} & 621.6G / 935K / \textbf{26.435} & 620.8G / 935K / \textbf{30.065} & \{v\} \\
            \cline{1-7}
        \end{tabular}
        }
    \end{subtable}
    \begin{subtable}[h]{0.275\linewidth}
        \caption{Reciprocal Helper (\textit{w/o} $/$ \textit{w/}).}
        \label{tab_ablation_helper}
        \centering
        \resizebox{\linewidth}{!}
        {
        \begin{tabular}{|l|c|c|}
        	\hline
            Task & Mult-Adds (G) & PSNR \\
            \hline
            SR $\times2$ & 163.2 / 163.4 & 35.308 / \textbf{35.324} \\
            SR $\times3$ & 77.16 / 77.26 & 31.482 / \textbf{31.508} \\
            SR $\times4$ & 42.08 / 42.13 & 29.308 / \textbf{29.374} \\
            LLE & 620.8 / 621.6 & 25.915 / \textbf{26.435} \\
            \hline
        \end{tabular}
        }
    \end{subtable}
    \begin{subtable}[h]{0.275\linewidth}
        \caption{MobiVari Activation Function.}
        \label{tab_ablation_act}
        \centering
        \resizebox{\linewidth}{!}
        {
        \begin{tabular}{|r|ccc|}
        	\hline
            Activation & SR $\times2$ & CDN $\sigma=50$ & LLE \\
            \hline
            ReLU6~\cite{howard2017mobilenets,sandler2018mobilenetv2} & 35.304 & 29.220 & 25.530 \\
            ReLU~\cite{nair2010rectified} & 35.322 & 29.270 & 26.160 \\
            GELU~\cite{hendrycks2016gaussian} & 35.320 & 29.268 & 26.230 \\
            Swish$_{\beta=1}$~\cite{ramachandran2017searching} & 35.306 & 29.250 & 26.160 \\
            \hline
            LeakyReLU~\cite{maas2013rectifier} & \textbf{35.324} & \textbf{29.275} & \textbf{26.435} \\
            \hline
        \end{tabular}
        }
    \end{subtable}
    \caption{Ablation studies on our proposed methods. The reported PSNR scores represent the average values on the benchmark test datasets of each image restoration task provided in Tabs.~\ref{tab_sr},~\ref{tab_drlle},~\ref{tab_dn}. Mult-Adds is calculated on a $1280\times720$ high-quality image.}
    \label{tab_ablation}
    \vspace{-10pt}
\end{table*}

\begin{table}[ht]
    \centering
    \begin{subtable}[h]{\linewidth}
        \caption{Color Image Denoising (CDN).}
        \label{tab_cdn}
        \centering
        \resizebox{0.95\linewidth}{!}
        {\begin{tabular}{|r|c|c|c|c|c|c|c|c|c|c|}
            \hline
            \multirow{2}{*}{Method} & \multirow{2}{*}{\#Params} & \multirow{2}{*}{$\sigma$} & \multicolumn{2}{c|}{CBSD68~\cite{martin2001database}} & \multicolumn{2}{c|}{Kodak24~\cite{franzen1999kodak}} & \multicolumn{2}{c|}{McMaster~\cite{zhang2011color}} & \multicolumn{2}{c|}{Urban100~\cite{huang2015single}}  \\ \cline{4-11}
            &&&PSNR&SSIM&PSNR&SSIM&PSNR&SSIM&PSNR&SSIM \\
            \hline
            SwinIR-light~\cite{liang2021swinir} & \textcolor{red}{905K} & \multirow{7}{*}{15} & \textcolor{blue}{34.16} & 0.9323 & \textcolor{blue}{35.18} & \textcolor{blue}{0.9269} & \textcolor{blue}{35.23} & \textcolor{blue}{0.9295} & \textcolor{blue}{34.59} & \textcolor{blue}{0.9478} \\
            Restormer-light~\cite{Zamir2021Restormer} & 1,054K & & 33.99 & 0.9311 & 34.86 & 0.9244 & 34.69 & 0.9229 & 34.00 & 0.9439 \\
            CAT-light~\cite{zheng2022cross} & 1,042K & & 34.01 & 0.9304 & 34.90 & 0.9237 & 34.83 & 0.9247 & 34.12 & 0.9443 \\
            ART-light~\cite{zhang2023accurate} & 1,084K & & 34.08 & 0.9315 & 35.00 & 0.9251 & 35.10 & 0.9282 & 34.44 & 0.9467 \\
            NGswin~\cite{choi2023n} & 993K & & 34.12 & \textcolor{blue}{0.9324} & 35.12 & 0.9268 & 35.17 & 0.9294 & 34.53 & 0.9476 \\
            \hhline{|=|=|~|=|=|=|=|=|=|=|=|}
            \textbf{RAMiT (ours)} & \textcolor{blue}{\textbf{935K}} & & \textcolor{red}{\textbf{34.23}} & \textcolor{red}{\textbf{0.9332}} & \textcolor{red}{\textbf{35.22}} & \textcolor{red}{\textbf{0.9276}} & \textcolor{red}{\textbf{35.31}} & \textcolor{red}{\textbf{0.9309}} & \textcolor{red}{\textbf{34.68}} & \textcolor{red}{\textbf{0.9488}} \\
            \hline\hline
            SwinIR-light~\cite{liang2021swinir} & \textcolor{red}{905K} & \multirow{7}{*}{25} & \textcolor{blue}{31.50} & 0.8883 & \textcolor{blue}{32.69} & \textcolor{blue}{0.8868} & \textcolor{blue}{32.90} & 0.8977 & \textcolor{blue}{32.23} & \textcolor{blue}{0.9222} \\
            Restormer-light~\cite{Zamir2021Restormer} & 1,054K & & 31.33 & 0.8865 & 32.38 & 0.8833 & 32.44 & 0.8905 & 31.60 & 0.9161 \\
            CAT-light~\cite{zheng2022cross} & 1,042K & & 31.37 & 0.8855 & 32.43 & 0.8822 & 32.58 & 0.8928 & 31.75 & 0.9167 \\
            ART-light~\cite{zhang2023accurate} & 1,084K & & 31.40 & 0.8864 & 32.49 & 0.8833 & 32.74 & 0.8956 & 32.03 & 0.9195 \\
            NGswin~\cite{choi2023n} & 993K & & 31.44 & \textcolor{blue}{0.8884} & 32.61 & 0.8865 & 32.82 & \textcolor{blue}{0.8978} & 32.13 & 0.9215 \\
            \hhline{|=|=|~|=|=|=|=|=|=|=|=|}
            \textbf{RAMiT (ours)} & \textcolor{blue}{\textbf{935K}} & & \textcolor{red}{\textbf{31.59}} & \textcolor{red}{\textbf{0.8902}} & \textcolor{red}{\textbf{32.76}} & \textcolor{red}{\textbf{0.8887}} & \textcolor{red}{\textbf{33.02}} & \textcolor{red}{\textbf{0.9008}} & \textcolor{red}{\textbf{32.36}} & \textcolor{red}{\textbf{0.9244}} \\
            \hline\hline
            SwinIR-light~\cite{liang2021swinir} & \textcolor{red}{905K} & \multirow{7}{*}{50} & \textcolor{blue}{28.22} & 0.8006 & \textcolor{blue}{29.54} & \textcolor{blue}{0.8089} & \textcolor{blue}{29.71} & \textcolor{blue}{0.8339} & \textcolor{blue}{28.89} & \textcolor{blue}{0.8658} \\
            Restormer-light~\cite{Zamir2021Restormer} & 1,054K & & 28.04 & 0.7974 & 29.19 & 0.8034 & 29.31 & 0.8256 & 28.30 & 0.8559 \\
            CAT-light~\cite{zheng2022cross} & 1,042K & & 28.11 & 0.7960 & 29.29 & 0.8024 & 29.48 & 0.8296 & 28.46 & 0.8573 \\
            ART-light~\cite{zhang2023accurate} & 1,084K & & 28.08 & 0.7950 & 29.27 & 0.8000 & 29.48 & 0.8279 & 28.62 & 0.8584 \\
            NGswin~\cite{choi2023n} & 993K & & 28.13 & \textcolor{blue}{0.8011} & 29.42 & 0.8087 & 29.59 & \textcolor{blue}{0.8339} & 28.75 & 0.8644 \\
            \hhline{|=|=|~|=|=|=|=|=|=|=|=|}
            \textbf{RAMiT (ours)} & \textcolor{blue}{\textbf{935K}} & & \textcolor{red}{\textbf{28.37}} & \textcolor{red}{\textbf{0.8058}} & \textcolor{red}{\textbf{29.67}} & \textcolor{red}{\textbf{0.8143}} & \textcolor{red}{\textbf{29.91}} & \textcolor{red}{\textbf{0.8422}} & \textcolor{red}{\textbf{29.15}} & \textcolor{red}{\textbf{0.8729}} \\
            \hline
        \end{tabular}}
    \end{subtable}
    \begin{subtable}[h]{0.95\linewidth}
        \caption{Grayscale Image Denoising (GDN).}
        \label{tab_gdn}
        \centering
        \resizebox{\linewidth}{!}
        {\begin{tabular}{|r|c|c|c|c|c|c|c|c|}
            \hline
            \multirow{2}{*}{Method} & \multirow{2}{*}{\#Params} & \multirow{2}{*}{$\sigma$} & \multicolumn{2}{c|}{Set12~\cite{zhang2017beyond}} & \multicolumn{2}{c|}{BSD68~\cite{martin2001database}} & \multicolumn{2}{c|}{Urban100~\cite{huang2015single}}  \\ \cline{4-9}
            &&&PSNR&SSIM&PSNR&SSIM&PSNR&SSIM \\
            \hline        
            SwinIR-light~\cite{liang2021swinir} & \textcolor{red}{903K} & \multirow{6}{*}{15} & \textcolor{blue}{33.04} & 0.9052 & 31.78 & 0.8926 & \textcolor{blue}{33.04} & \textcolor{blue}{0.9317} \\
            Restormer-light~\cite{Zamir2021Restormer} & 1,053K & & 32.93 & 0.9039 & 31.76 & 0.8922 & 32.81 & 0.9306 \\
            CAT-light~\cite{zheng2022cross} & 1,041K & & 32.91 & 0.9021 & \textcolor{red}{31.89} & 0.8913 & 31.80 & 0.8901 \\
            ART-light~\cite{zhang2023accurate} & 1,082K & & 32.93 & 0.9023 & 31.73 & 0.8911 & 32.89 & 0.9299 \\
            NGswin~\cite{choi2023n} & 991K & & \textcolor{blue}{33.04} & \textcolor{blue}{0.9055} & 31.78 & \textcolor{blue}{0.8927} & 32.99 & 0.9314 \\
            \hhline{|=|=|~|=|=|=|=|=|=|}
            \textbf{RAMiT (ours)} & \textcolor{blue}{\textbf{932K}} & & \textcolor{red}{\textbf{33.14}} & \textcolor{red}{\textbf{0.9070}} & \textcolor{blue}{\textbf{31.82}} & \textcolor{red}{\textbf{0.8939}} & \textcolor{red}{\textbf{33.19}} & \textcolor{red}{\textbf{0.9346}} \\
            \hline\hline
            SwinIR-light~\cite{liang2021swinir} & \textcolor{red}{903K} & \multirow{6}{*}{25} & \textcolor{blue}{30.67} & 0.8669 & 29.32 & 0.8325 & \textcolor{blue}{30.52} & \textcolor{blue}{0.8963} \\
            Restormer-light~\cite{Zamir2021Restormer} & 1,053K & & 30.60 & 0.8659 & 29.32 & 0.8322 & 30.32 & 0.8952 \\
            CAT-light~\cite{zheng2022cross} & 1,041K & & 30.60 & 0.8641 & \textcolor{red}{29.47} & \textcolor{blue}{0.8330} & 29.32 & 0.8393 \\
            ART-light~\cite{zhang2023accurate} & 1,082K & & 30.52 & 0.8620 & 29.25 & 0.8285 & 30.30 & 0.8919 \\
            NGswin~\cite{choi2023n} & 991K & & 30.65 & \textcolor{blue}{0.8671} & 29.33 & 0.8324 & 30.46 & 0.8961 \\
            \hhline{|=|=|~|=|=|=|=|=|=|}
            \textbf{RAMiT (ours)} & \textcolor{blue}{\textbf{932K}} & & \textcolor{red}{\textbf{30.79}} & \textcolor{red}{\textbf{0.8694}} & \textcolor{blue}{\textbf{29.37}} & \textcolor{red}{\textbf{0.8346}} & \textcolor{red}{\textbf{30.71}} & \textcolor{red}{\textbf{0.9013}} \\
            \hline\hline
            SwinIR-light~\cite{liang2021swinir} & \textcolor{red}{903K} & \multirow{6}{*}{50} & \textcolor{blue}{27.50} & \textcolor{blue}{0.7966} & 26.35 & \textcolor{blue}{0.7299} & \textcolor{blue}{27.01} & 0.8190 \\
            Restormer-light~\cite{Zamir2021Restormer} & 1,053K & & 27.48 & 0.7960 & 26.38 & 0.7285 & 26.92 & 0.8190 \\
            CAT-light~\cite{zheng2022cross} & 1,041K & & 27.49 & 0.7935 & \textcolor{red}{26.52} & \textcolor{red}{0.7333} & 26.06 & 0.7456 \\
            ART-light~\cite{zhang2023accurate} & 1,082K & & 27.26 & 0.7856 & 26.25 & 0.7194 & 26.68 & 0.8065 \\
            NGswin~\cite{choi2023n} & 991K & & 27.42 & 0.7961 & 26.38 & 0.7298 & 26.96 & \textcolor{blue}{0.8192} \\
            \hhline{|=|=|~|=|=|=|=|=|=|}
            \textbf{RAMiT (ours)} & \textcolor{blue}{\textbf{932K}} & & \textcolor{red}{\textbf{27.65}} & \textcolor{red}{\textbf{0.8013}} & \textcolor{blue}{\textbf{26.46}} & \textcolor{red}{\textbf{0.7333}} & \textcolor{red}{\textbf{27.32}} & \textcolor{red}{\textbf{0.8306}} \\
            \hline
        \end{tabular}}
    \end{subtable}
    \caption{Comparison of lightweight blind image denoising results. We refer to the baselines in~\cite{choi2023exploration}.}
    \label{tab_dn}
    \vspace{-15pt}
\end{table}

\textbf{Image Super-Resolution (SR).}
In Tab.~\ref{tab_sr}, we compared our RAMiT with other state-of-the-art lightweight SR methods, including CARN (ECCV18)~\cite{ahn2018fast}, LatticeNet (ECCV20)~\cite{luo2020latticenet}, SwinIR-light (ICCVW21)~\cite{liang2021swinir}, FMEN (CVPRW22)~\cite{du2022fast}, ESRT (CVPRW22)~\cite{Lu_2022_CVPR}, ELAN-light (ECCV22)~\cite{zhang2022efficient}, DiVANet (PR23)~\cite{behjati2023single}, NGswin (CVPR23)~\cite{choi2023n}, and SwinIR-NG (CVPR23)~\cite{choi2023n}.
We also reported the number of operations (Mult-Adds) of each model.
Our RAMiT gained PSNR up to 0.12dB while consuming only $59.6 \sim 67.7\%$ of the operations used by SwinIR-NG.
Especially, RAMiT offers the best trade-off between efficiency and performance on $\times2$ and $\times4$ tasks among the compared approaches.
For a concern of the number of parameters, see Appendix Sec.~\ref{appendix_sr}.

\textbf{Low-Light Image Enhancement (LLE).}
RAMiT substantially surpassed the state-of-the-art LLE methods, including DRBN (CVPR20)~\cite{yang2020fidelity}, KinD++ (IJCV21)~\cite{zhang2021beyond}, EnlightenGAN (TIP21)~\cite{jiang2021enlightengan}, and URetinex-Net (CVPR22)~\cite{wu2022uretinex}, as recorded in Tab.~\ref{tab_lle}.
Our method restored much more accurate brightness and objects from the extremely dark image than other models.
While they adhered to the conventional approaches, such as Retinex algorithms~\cite{rahman2004retinex} and convolutional neural networks, our advanced Transformer easily defeated them by up to 6.52dB of the PSNR score.

\textbf{Image Deraining (DR).}
Tab.~\ref{tab_dr} shows that RAMiT could more sufficiently remove rains than the state-of-the-art DR methods: UMRL (CVPR19)~\cite{yasarla2019uncertainty}, MSPFN (CVPR20)~\cite{jiang2020multi}, DRT (CVPRW22)~\cite{liang2022drt}, and TAO-Net (SPLetters22)~\cite{li2022tao}.
We gained PSNR scores up to 1.73dB with the second smallest architecture.
In particular, MSPFN network fell behind RAMiT in performance despite having 3.89 times more parameters than RAMiT. 

\textbf{Color Image Denoising (CDN).}
In Tab.~\ref{tab_cdn}, we referred to the lightweight denoising Transformer baselines introduced by~\cite{choi2023exploration}, such as SwinIR-light (ICCVW21)~\cite{liang2021swinir}, Restormer-light (CVPR22)~\cite{Zamir2021Restormer}, CAT-light (NeurIPS22)~\cite{zheng2022cross}, ART-light (ICLR23)~\cite{zhang2023accurate}, and NGswin (CVPR23)~\cite{choi2023n}.
It is notable that SwinIR, CAT, and NGswin aimed to boost locality of a window-based spatial self-attention, while Restoremer and ART pursued an improved ability in capturing non-local dependency in an image.
However, RAMiT surpassed them on every noise level and dataset through both local and global context.

\textbf{Grayscale Image Denoising (GDN).}
As shown in Tab.~\ref{tab_gdn}, our RAMiT was good at removing noise from the grayscale images as well.
RAMiT reconstructed more similar images to ground-truth for human-perception in that our SSIM scores were the highest.
Moreover, RAMiT gained PSNR scores on all noise levels up to 0.23dB.

\subsection{Ablation Study}
\label{ablation}

\textbf{D-RAMiT.}
Tab.~\ref{tab_ablation0} (\{i\} \emph{vs.} \{iii\} \emph{vs.} \{v\}) compares our D-RAMiT with a pure SPSA and CHSA on SR $\times2$, $\times4$, CDN, LLE, and DR.
The proposed D-RAMiT overcame the limited capacity of CHSA and the narrow receptive field of SPSA.
Our method achieved better results on multiple tasks with fewer computations and parameters than SPSA.
This effectiveness is also observed without the H-RAMi layer, another proposed method (\{ii\} \emph{vs.} \{iv\}).
Moreover, as shown in Tab.~\ref{tab_ablation_helper}, the reciprocal helper contributed to the improvement.
This approach consumed only minor amounts of Mult-Adds and no extra parameters.
Therefore, it was proven that our dimensional reciprocal self-attention mixing Transformers could be suitable for general IR tasks.

\textbf{H-RAMi.}
Tab.~\ref{tab_ablation0} (\{ii\} \emph{vs.} \{iii\}, \{iv\} \emph{vs.} \{v\}) revealed that H-RAMi constituted another critical component, not only for our D-RAMiT but also for a pure SPSA.
Regardless of tasks, this layer enabled the models to remain robust even when a hierarchical network caused information losses.
We assumed that since a noisy image contained more distorted boundaries, the impacts of H-RAMi that could recover more accurate object boundaries (Sec.~\ref{hrami}) were particularly significant in denoising tasks.
Additionally, the results highlighted the remarkable efficiency in that H-RAMi required marginal additional operations and parameters, which accounted for a maximum of only 3.01\% and 2.25\% of the total costs, respectively.

\textbf{MobiVari.}
In Tab.~\ref{tab_ablation_act}, we investigated different non-linear activation functions for our MobiVari.
LeakyReLU~\cite{maas2013rectifier} resulted in the best stable performances across multiple tasks and was selected as the default option.
Such stability of LeakyReLU can be attributed to its ability to better preserve relatively large absolute negative values compared to other activation functions. 
These values, which are occasionally generated by intermediate layers, may have a subtle influence on a feature map, ultimately leading to a significant difference in the final output of a network.
\section{Conclusion}
\label{conclusion}
This paper proposed the Reciprocal Attention Mixing Transformers (RAMiT).
To incorporate local and global context in an image, our Dimensional Reciprocal Attention Mixing Transformer (D-RAMiT) blocks computed bi-dimensional self-attentions in parallel and mixed them.
The reciprocal helper was useful for this mechanism.
Moreover, the Hierarchical Reciprocal Attention Mixing (H-RAMi) layer was also introduced, where the information losses caused by downsampling were complemented.
For mixing attentions and other convolutional layers, we revisited and modified the MobileNet.
As a result, our RAMiT achieved state-of-the-art performances on multiple lightweight image restoration tasks, including super-resolution, low-light enhancement, deraininig, color denoising, and grayscale denoising.
In closing, we hope this work can be further developed and extended to other low-level tasks.

\hspace{-12pt}{\bf Acknowledgements.} This paper was supported by Institute of Information \& Communications Technology Planning \& Evaluation (IITP) grant (No.2022-0-00956) and Korea Health Industry Development Institute (KHIDI) grant (No. H122C1983) funded by the Korea government (MSIT). Special thanks to Prof. Dr. Dasaem Jeong for valuable advises, and Namo Bang, Jeongkyun Park, Juntae Kim, and Gyusik Choi of the Sogang Synergy Journal Club for precious comments.

\setcounter{section}{0}
\setcounter{table}{0}
\setcounter{figure}{0}
\renewcommand\thesection{\Alph{section}}
\renewcommand\thetable{\Alph{table}}
\renewcommand\thefigure{\Alph{figure}}

\vspace{10pt}
{\Large\bf {Appendix}}
\section{Supplementary Discussions and Ablation Studies}
\label{appendix_discussion_ablation}
\subsection{MobiVari (MobileNet Variants) and Reconstruction Module}
\label{appendix_mobivari_reconstruction}

\begin{figure}[h]
\centering
\includegraphics[width=\linewidth]{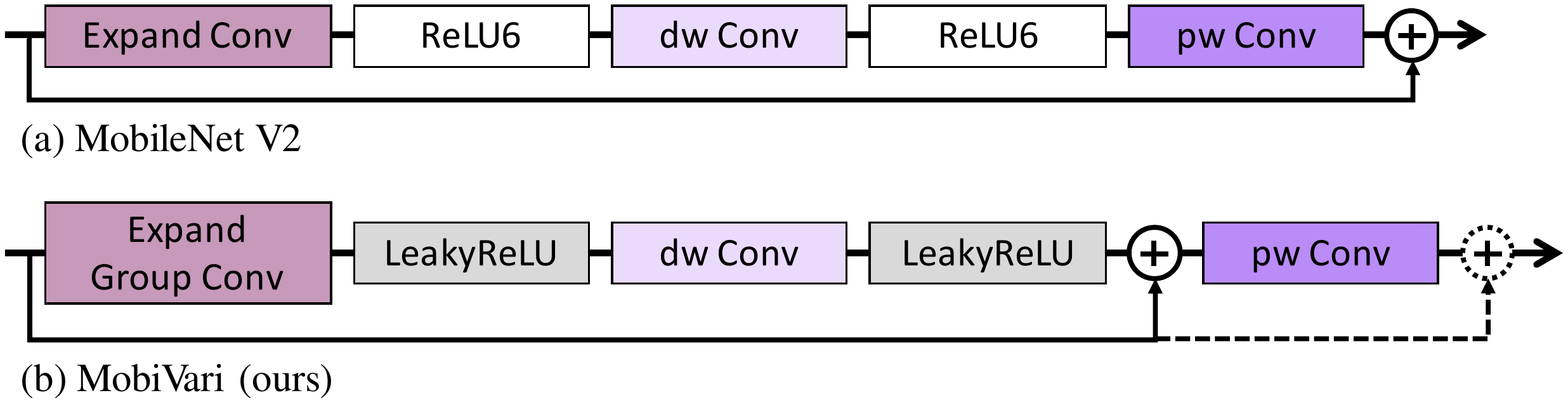}
\caption{Comparison of MobileNets V2 and our corresponding variants, MobiVari.}
\label{appendix_fig_mobivari}
\end{figure}

We revisit MobileNet V2 architectures~\cite{sandler2018mobilenetv2} to incorporate simple and efficient CNN structures into our components.
Fig.~\ref{appendix_fig_mobivari} illustrates a comparison between the MobileNet and our modified version.
We replace the ReLU6 non-linearity~\cite{sandler2018mobilenetv2} with LeakyReLU~\cite{maas2013rectifier} to preserve subtle gradients that ReLU6 cannot capture~\cite{maas2013rectifier}.
Empirical evidence in Tab.~\ref{tab_ablation_act} of the main paper shows that this change is the most stable.
The $3\times3$ depth-wise (\textit{dw}) and $1\times1$ point-wise (\textit{pw}) convolutions in MobileNets are residually connected~\cite{he2016deep} with the input feature.
However, if the channels produced by \textit{pw} convolutions differ from input channels, the skip connection for \textit{pw} convolutions is ignored.
Furthermore, because the first $1\times1$ convolution expanding channels in MobileNet V2 requires many parameters and computations, it is not suitable for our lightweight design.
Therefore, we substitute it with group convolution~\cite{cohen2016group}, where the group size and expansion ratio are set to 4 and 1.2, respectively, by default.
Our MobiVari is applied to attention mixing layers of D-RAMiT and H-RAMi, a downsizing layer, a bottleneck, and the reconstruction module.

\begin{figure}[t]
    \centering
    \begin{subfigure}{\linewidth}
        \centering
        \includegraphics[width=0.7\linewidth]{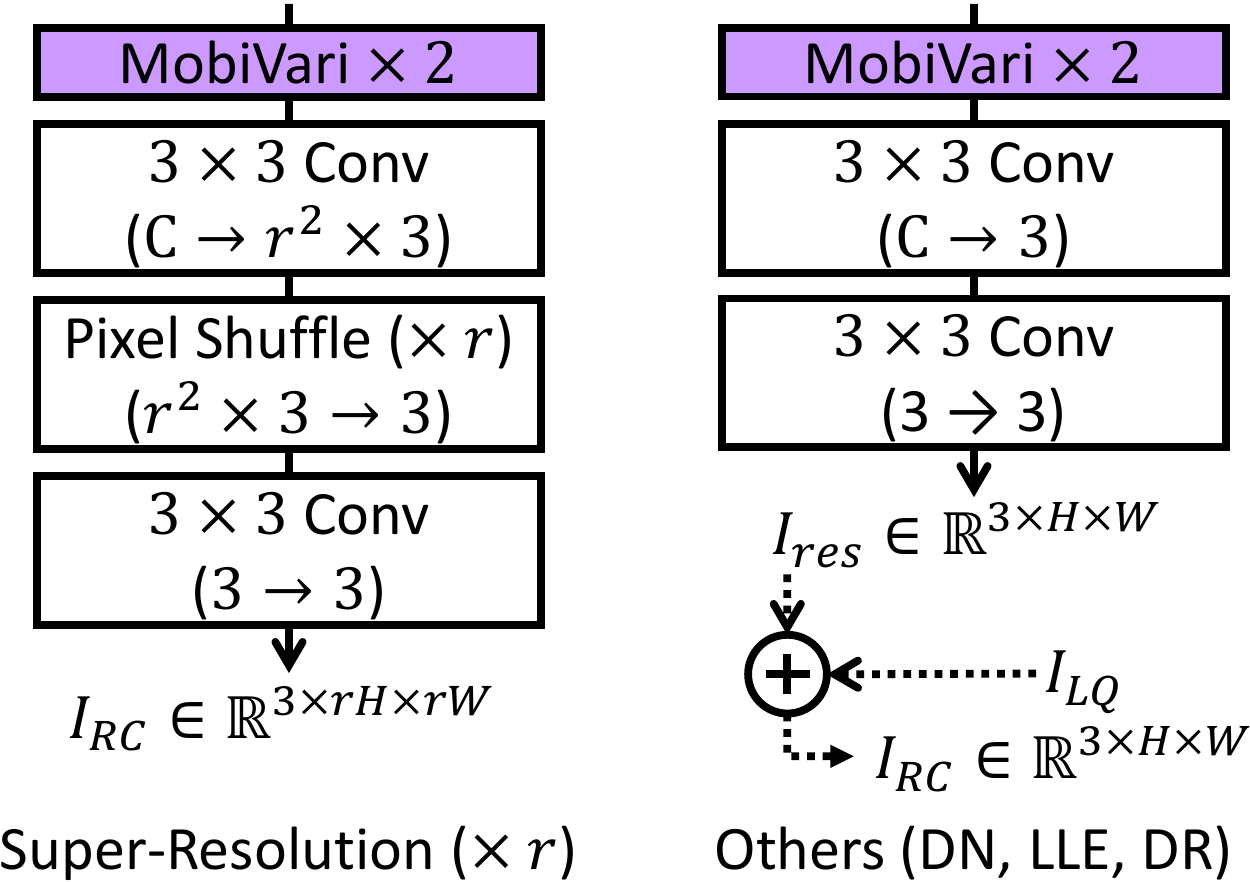}
        \caption{Reconstruction module architecture.}
        \label{appendix_fig_reconstruction_arch}
    \end{subfigure}
    \begin{subfigure}{\linewidth}
        \centering
        \includegraphics[width=0.85\linewidth]{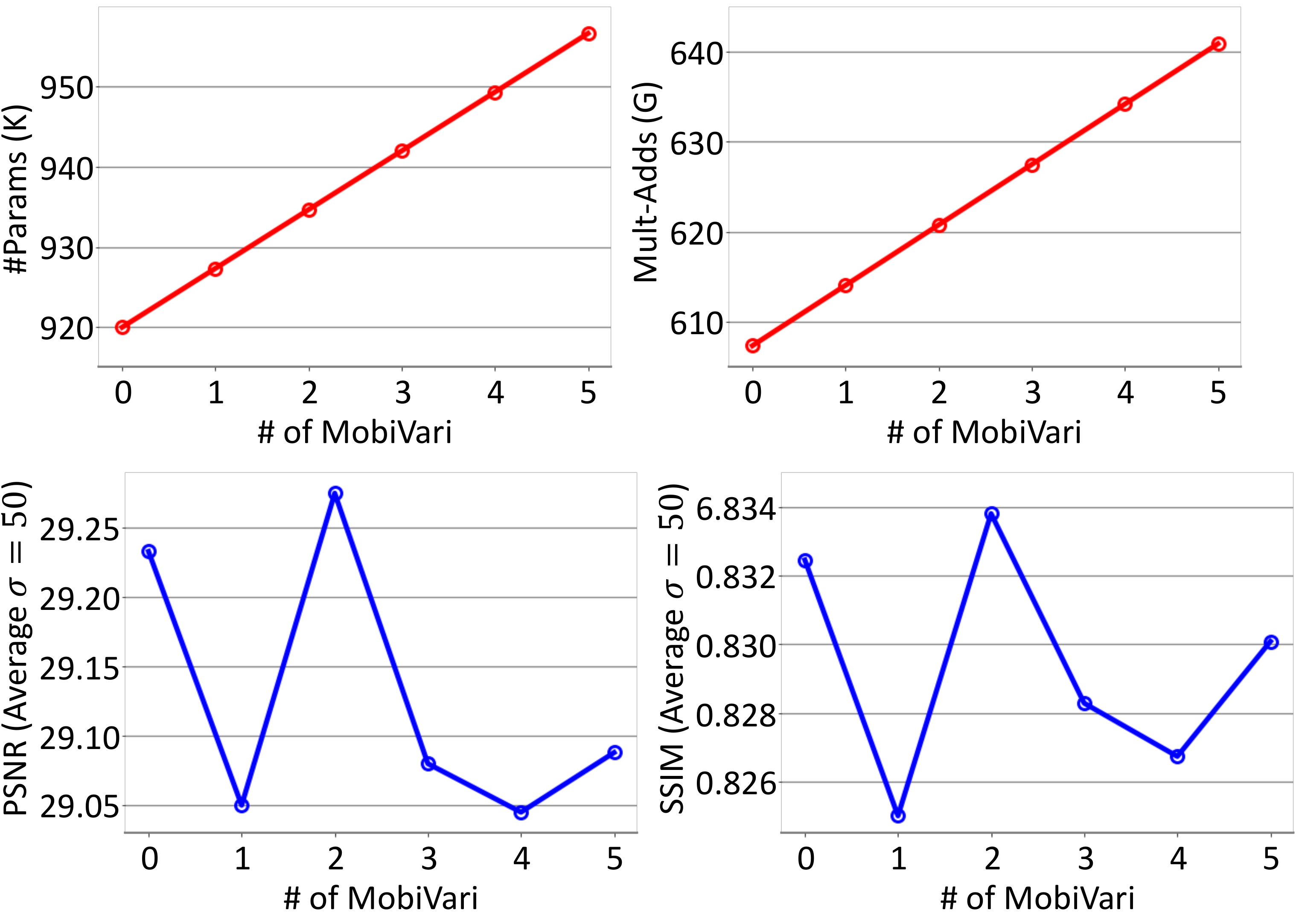}
        \caption{Ablation study on the number of MobiVari layers at the reconstruction module. The metrics are evaluated on color denoising task using $\sigma=50$. PSNR and SSIM average the scores on four benchmark datasets.}
        \label{appendix_fig_tailmv}
    \end{subfigure}
    \caption{Reconstruction module.}
    \label{appendix_fig_reconstruction}
\end{figure}

Fig.~\ref{appendix_fig_reconstruction_arch} depicts the reconstruction module (a final layer).
The basic structure follows the reconstruction module of NGswin~\cite{choi2023n}.
The only difference is that we place two MobiVari layers before the default version to balance the trade-off between performance and efficiency (See Fig.~\ref{appendix_fig_tailmv}).
This module slightly varies depending on tasks.
For super-resolution, a pixel-shuffler~\cite{shi2016real} is employed to upscale the feature maps by $r$ times.
However, since other tasks (denoising, low-light enhancement, and deraining) do not require this process, the pixel-shuffler is discarded.
The symbols and numbers in parentheses indicate changes of channels.
The operation $I_{res}+I_{LQ}$ follows convention~\cite{liang2021swinir,Zamir2021Restormer}.

\subsection{Bi-dimensional Self-Attention}
\label{appendix_davit}

\begin{table*}[t]
    \centering
    \begin{subtable}[h]{\linewidth}
        \caption{Attribute comparisons. The text in \textbf{bold} indicates the key differences. ``LN'' represents whether the position of layer-norm~\cite{ba2016layer} is before (Pre) or after (Post) the self-attention and feed-forward network.}
        \label{appendix_tab_davit_attr}
        \centering
        \resizebox{0.9\linewidth}{!}
        {
        \begin{tabular}{|r|ll|rrrr|r|}
            \hline
            \multirow{2}{*}{Method} & \multicolumn{2}{|c|}{\textbf{SPSA \& CHSA}} & \multicolumn{4}{|c|}{Existing Elements Employed} & \multirow{2}{*}{Solving Problems} \\
            \cline{2-7}
            & \textbf{Operating} & \textbf{Importance on} & Window Shift & Positional Encoding & Self-Attention & LN & \\
            \hline
            DaViT~\cite{ding2022davit} & \textbf{Alternatively} & \textbf{Both equally} & No use & Convolution~\cite{Islam*2020How} & Scaled dot-product~\cite{vaswani2017attention} & Pre~\cite{dosovitskiy2021an} & High-level vision \\
            D-RAMiT (ours) & \textbf{In parallel} & \textbf{SPSA more} & Cyclic~\cite{liu2021swin} & Relative Position Bias~\cite{shaw-etal-2018-self,liu2021swin} & Scaled cosine~\cite{liu2022swin} & Post~\cite{liu2022swin} & Low-level vision \\
            \hline
        \end{tabular}
        }
    \end{subtable}
    \begin{subtable}[h]{\linewidth}
        \vspace{3pt}
        \caption{Ablation study on DaViT (Mult-Adds / \#Params / Average PSNR).}
        \label{appendix_tab_ablation_davit}
        \centering
        \resizebox{0.9\linewidth}{!}
        {
        \begin{tabular}{|r|c|c|c|c|c|}
            \hline
            Method & SR $\times2$ & SR $\times4$ & CDN $\sigma=50$ & LLE & DR \\
            \hline
            DaViT-\textit{full}~\cite{ding2022davit} & 167.0G / 983K / 35.064 & 43.0G / 1,003K / 29.088 & 635.2G / 977K / 28.785 & 635.2G / 977K / 20.965 & 635.2G / 977K / 27.910 \\
            DaViT-\textit{core}~\cite{ding2022davit} & 163.2G / 966K / 35.172 & 42.08G / 987K / 29.268 & 620.8G / 961K / 29.108 & 620.8G / 961K / 26.000 & 620.8G / 961K / 29.630 \\
            \textbf{RAMiT (ours)} & \textbf{163.4G} / \textbf{\textcolor{red}{940K}} / \textbf{\textcolor{red}{35.324}} & \textbf{42.13G} / \textbf{\textcolor{red}{961K}} / \textbf{\textcolor{red}{29.374}} & \textbf{620.8G} / \textbf{\textcolor{red}{935K}} / \textbf{\textcolor{red}{29.275}} & \textbf{621.6G} / \textbf{\textcolor{red}{935K}} / \textbf{\textcolor{red}{26.435}} & \textbf{620.8G} / \textbf{\textcolor{red}{935K}} / \textbf{\textcolor{red}{30.065}} \\
            \hline
        \end{tabular}
        }
    \end{subtable}
    \caption{Comparisons of RAMiT and DaViT~\cite{ding2022davit}.}
    \label{appendix_tab_davit}
\end{table*}

\begin{figure}[t]
    \centering
    \includegraphics[width=\linewidth]{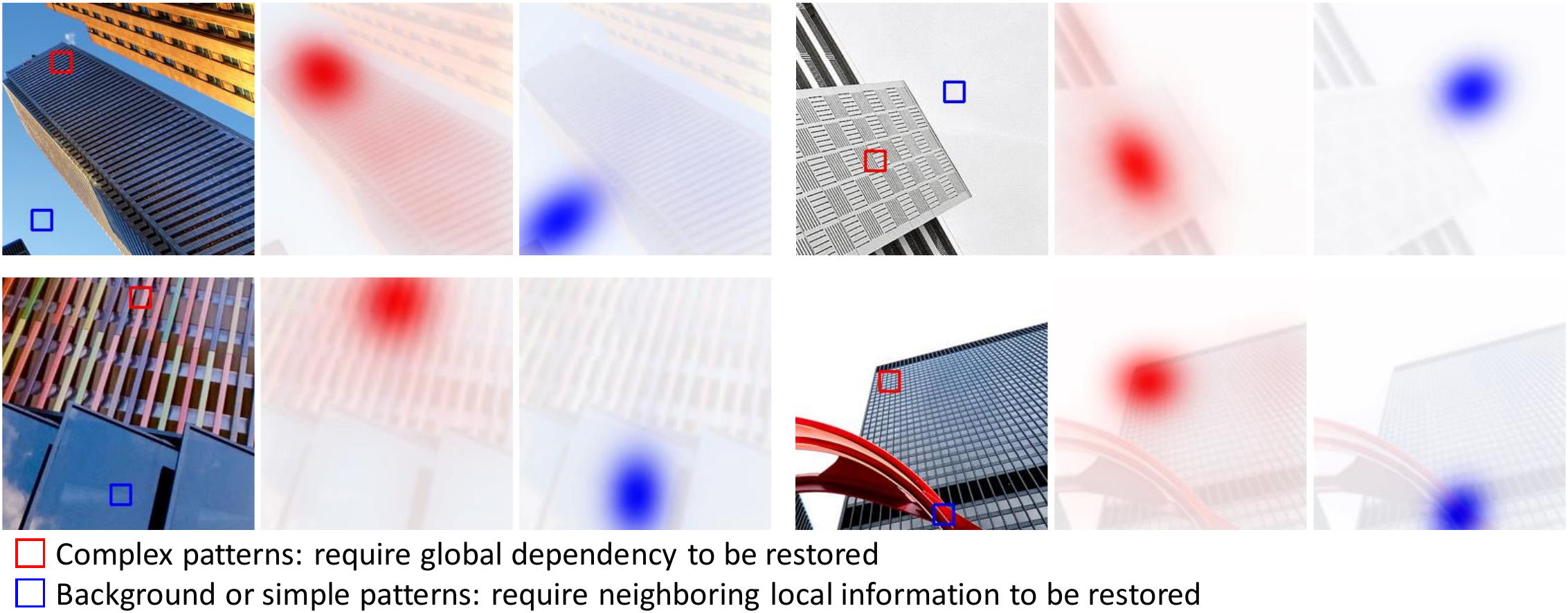}
    \caption{The importance of capturing both local and global context for restoring different parts.}
    \label{appendix_fig_localglobal}
\end{figure}

Regarding the importance of capturing both local and global context, we present Fig.~\ref{appendix_fig_localglobal}.
In this figure, while complex patterns in the image require global context to recover, background or simple patterns require only neighboring local information.

Similar to our bi-dimensional self-attention of the proposed D-RAMiT blocks, DaViT~\cite{ding2022davit} also developed a Transformer using both spatial self-attention (SPSA) and channel self-attention (CHSA).
Tab.~\ref{appendix_tab_davit_attr} summarizes the attributes of DaViT and D-RAMiT.
A core difference is that DaViT ``alternatively'' places SPSA and CHSA, while D-RAMiT operates them ``in parallel''.
As discussed in Sec.~\ref{drami} of the main text, our architecture can boost (depending on tasks) both SPSA and CHSA through the reciprocal helper, which DaViT fundamentally cannot utilize.
Another crucial distinction is related to ``which self-attention module is given more importance''.
While D-RAMiT assigns more multi-heads on SPSA, DaViT makes the number of both modules identical. 
We hypothesize that DaViT's simple approach is unsuitable for lightweight image restoration because although CHSA can capture global dependency, its performance is significantly impaired under parameter constraints, as observed in Tab.~\ref{tab_ablation0} of our main body.
Therefore, more weights on SPSA can be more useful for constructing an effective lightweight RAMiT. Other differences are summarized in the table.

To further demonstrate our superiority over the simple bi-dimensional approach of DaViT, we constructed two versions in Tab.~\ref{appendix_tab_ablation_davit}.
The first version, DaViT-\textit{full}, replaced the D-RAMiT blocks' elements in Tab.~\ref{appendix_tab_davit_attr} with those of DaViT.
The second version, DaViT-\textit{core}, changed only the core designs (\emph{i.e.}, SPSA \& CHSA ``Operating'' and ``Importance on'') from ours to those of DaViT, while the parts of the ``Existing Elements Employed'' column remained as our settings.
The other elements not mentioned in the table followed our default settings for a fair comparison, including the shallow module, MobiVari, the downsizing layers, the bottleneck layer, H-RAMi layer, the reconstruction module, and the hyper-parameters of Tab.~\ref{appendix_tab_implement} (except that \texttt{chsa\_head\_ratio} is no longer needed).
The results show that RAMiT outperformed both DaViT versions while having fewer parameters and almost the same Mult-Adds.
It is demonstrated that our meticulous composition of SPSA and CHSA can make a significant difference for multiple lightweight image restoration tasks.
\subsection{LAM Comparisons with Other Models}
\label{appendix_lam}

\begin{figure}[t]
\centering
\includegraphics[width=\linewidth]{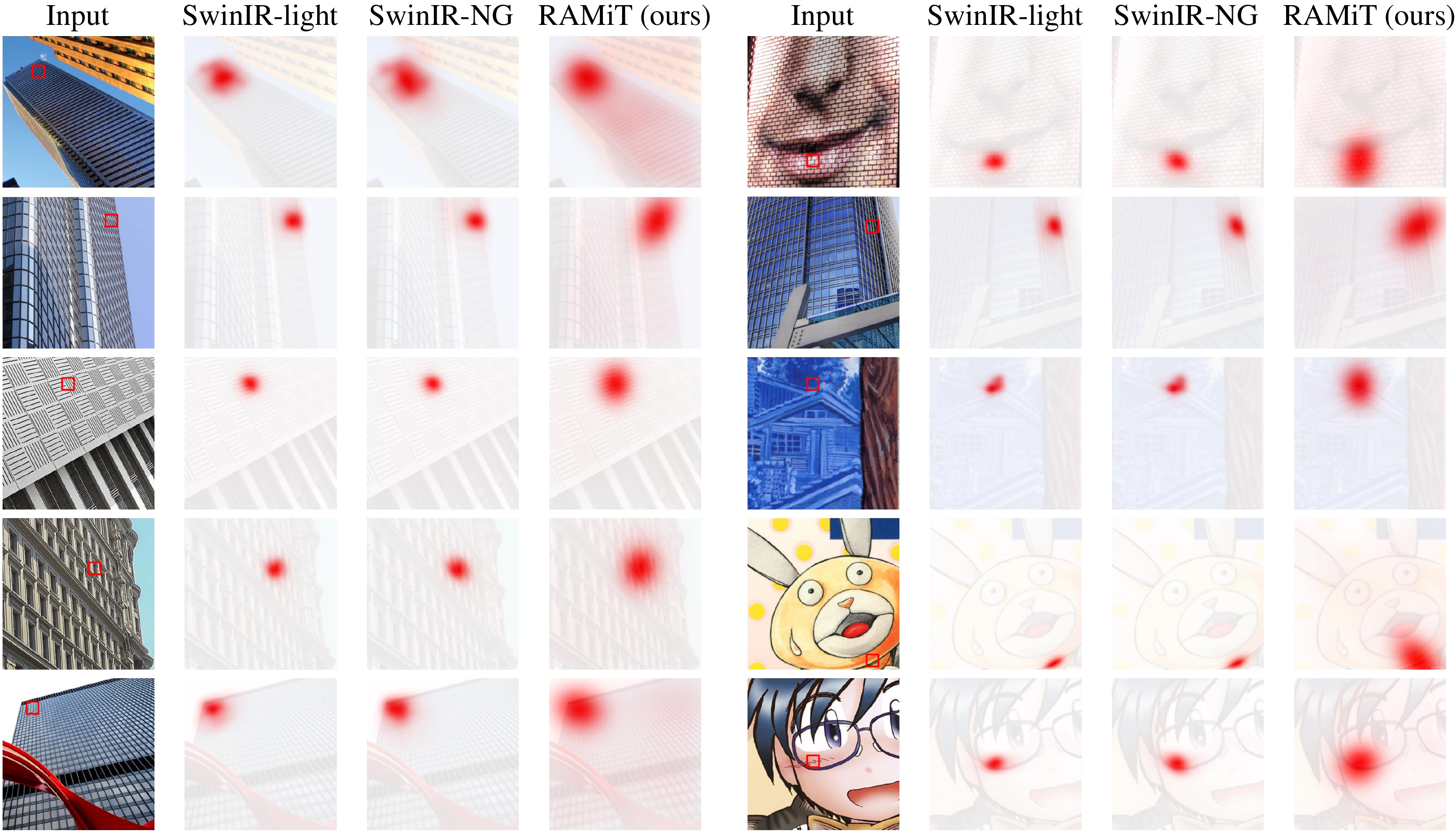}
\caption{Local Attribution Map (LAM)~\cite{gu2021interpreting} comparison. The depth of the red areas indicates the extent to which the regions contribute to recovering a \textcolor{red}{red} box of an input.}
\label{appendix_fig_lam}
\end{figure}

SwinIR-light (ICCVW21)~\cite{liang2021swinir} is the first successful attempt applying window self-attention (WSA) to the image restoration tasks.
Most recently, SwinIR-NG (CVPR23)~\cite{choi2023n} defined an N-Gram context method enlarging the regions viewed for recovering distorted pixels, to solve the limited ``local'' receptive field problem of SwinIR-light.
However, SwinIR-NG failed to capture ``global context'', while our RAMiT successfully exploit the ``global receptive field'' maintaining WSA approach, which is clarified by LAM~\cite{gu2021interpreting} results in Fig.~\ref{appendix_fig_lam}.
Even if SwinIR-NG tends to utilize the slightly expanded receptive field when compared to SwinIR-light, the gradients of SwinIR-NG that actually contribute to reconstruct a small red box are limited within ``local areas''.
By contrast, our RAMiT can convey the gradients to ``global regions'', which improves low-level vision performances with fewer computational costs than SwinIR-NG (reference Tab.~\ref{tab_sr} of the main paper).

This ability results from adoption of channel self-attention.
According to prior work, Squeeze-and-Excitation networks~\cite{hu2018squeeze}, the channel-attention can effectively embed the ``global feature responses''.
RCAN~\cite{zhang2018image} delivered an insight that channel-wise attention would be good at modeling ``global spatial dependency'' for low-level vision tasks.
Afterwards, Restormer~\cite{Zamir2021Restormer} applied this mechanism to self-attention without squeeze operations, thereby preserving abundant spatial information, which enabled the image restoration networks to more effectively capture the ``global interdependencies'' in a whole image.
Exploiting such advantages of channel self-attention and the effective WSA, RAMiT can yield meaningfully larger receptive fields than the ``pure local-attention'' of the SwinIR family.
Therefore, our work can be considered an enhanced version of the N-Gram context~\cite{choi2023n}, which extends the ``local'' N-Gram approach to a ``Global-Gram'' method.
\subsection{Reciprocal Helper}
\label{appendix_helper}

\begin{table}[h]
    \centering
    \resizebox{0.85\linewidth}{!}
    {
    \begin{tabular}{|l|c|c|}
        \hline
        Task & Mult-Adds (G) & PSNR \\
        \hline
        Color Denoising & 620.8 / 621.6 & \textbf{29.275} / 29.253 \\
        Grayscale Denoising & 618.5 / 619.3 & \textbf{27.143} / 27.100 \\
        Deraining & 620.8 / 621.6 & \textbf{30.065} / 29.960 \\
        \hline
    \end{tabular}
    }
    \caption{Ablation study on the proposed Reciprocal Helper for denoising and deraining (\textit{w/o} / \textit{w/}).}
    \label{appendix_tab_ablation_helper}
\end{table}

As proved in Tab.~\ref{tab_ablation_helper} of the main content, our Reciprocal Helper\footnote{To prevent any confusion, we adopted the term ``\textit{Dimensional Reciprocal Attention Mixing Transformer}'' (D-RAMiT) to indicate that \textit{every dimension} (spatial and channel) of feature maps is utilized in calculating \textit{self-attention}, and the outcomes are subsequently \textit{mixed} by MobiVari. Consequently, this implies that the reciprocal helper is not a prerequisite to represent dimensional reciprocal attention.} can boost $\times2,\times3,\times4$ super-resolution and low-light enhancement tasks.
However, Tab.~\ref{appendix_tab_ablation_helper} shows that this mechanism is unable to improve the performances of denoising and deraining.
We interpret this limitation in terms of properties of the tasks.
Degradation used for the super-resolution and low-light enhancement inputs relatively has regularity and therefore may be easy to be globally encoded.
This property may make our reciprocal helper useful for the parallel process of local and global self-attention.
On the other hand, when dealing with denoising or deraining low-quality inputs, the network is required to erase somethings that obscure the high-quality objects or background.
Since it is ill-posed to globally encode these (randomly) disorganized obstructions with a small network capacity, the global embeddings produced by the channel attention may confuse the spatial attention module of the next blocks.
However, if the parallel process lacks the reciprocal helper, the MobiVari mixing layers alone can still resolve this issue well.
Admitting this limitation, we will conduct more sophisticated future work on other helper algorithms that can improve universal tasks.
Nevertheless, our core ideas, \emph{i.e.,} dimensional and hierarchical reciprocal self-attention methods, have been already demonstrated to be effective and efficient enough to achieve new state-of-the-art lightweight denoising and deraining.
\subsection{Hierarchical Reciprocal Attention Mixing Layer (H-RAMi)}
\label{appendix_hrami}

\begin{figure}[t]
\centering
\includegraphics[width=\linewidth]{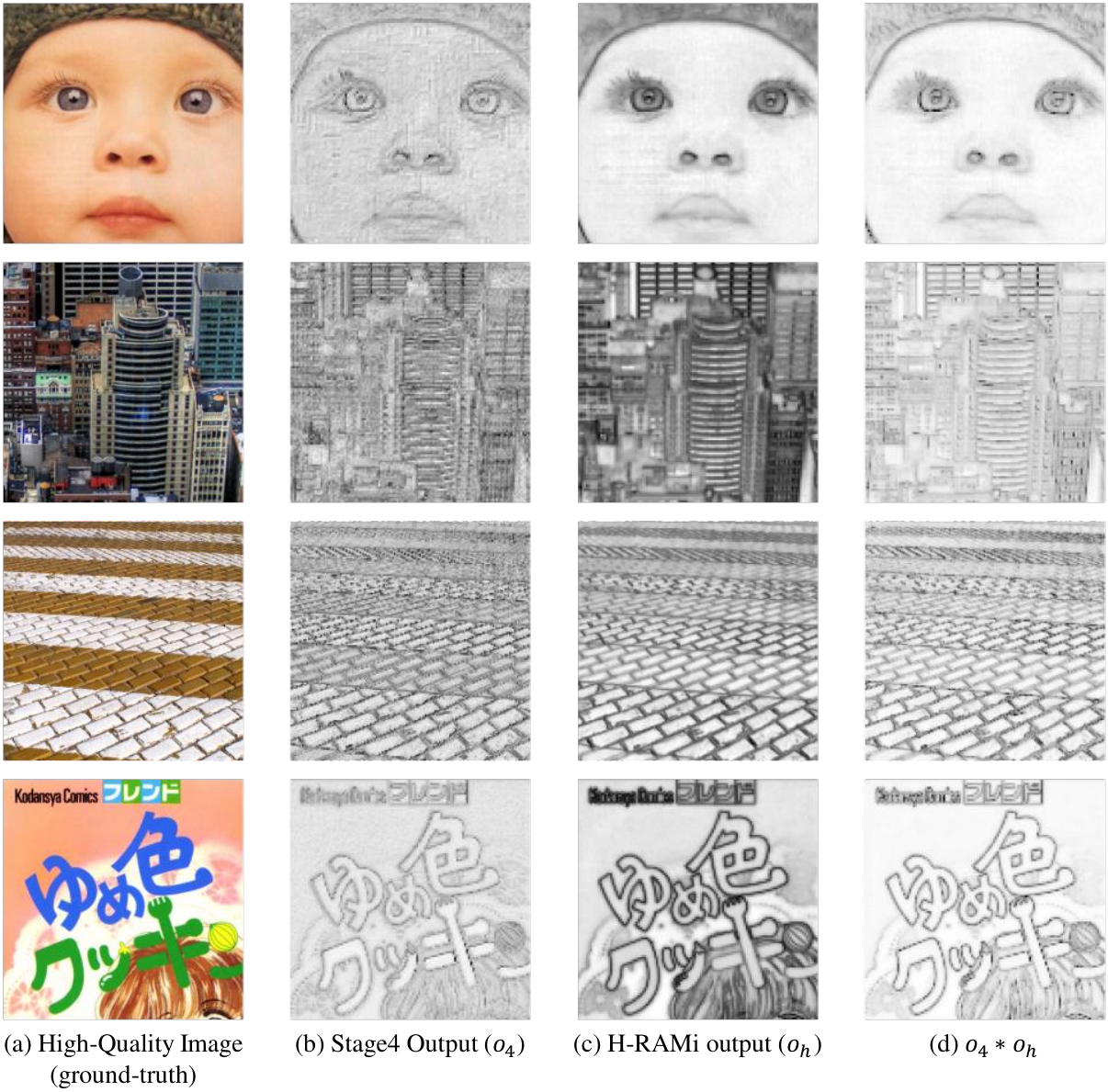}
\caption{Impacts of H-RAMi. \textbf{(a)} A ground-truth high-quality image. \textbf{(b)}, \textbf{(c)} The feature maps after stage $4$ and H-RAMi. \textbf{(d)} Element-wise product of (b) and (c). (b), (c), (d) are obtained by max-pooling along channel and standardization.}
\label{appendix_fig_hrami}
\end{figure}

Although H-RAMi may appear similar to the attention banks used in DiVANet~\cite{behjati2023single}, there are notable differences.
DiVANet uses non-hierarchical attentions for every residual convolution block, increasing computational costs (see Tab.~\ref{appendix_tab_slimsr}) and failing to learn semantic-level representation.
Moreover, the vertical and horizontal squeeze operations prevent the attention layers from considering full-resolution information.
In contrast, our approach reduces time complexity and utilizes semantic-level information by processing compressed feature maps.
Furthermore, the inputs to H-RAMi are intermediate attentions from D-RAMiT blocks, which preserve information from both full-resolution spatial and channel self-attentions.
We provide additional visual evidences of the benefits in Fig.~\ref{appendix_fig_hrami}.
As previously stated in Fig.~\ref{fig_hrami} of the main text, the stage $4$ output alone at (b) produces relatively unclear or incorrect edges, which are resolved at (d) by the clearer edges produced by H-RAMi at (c).
\subsection{Super-Resolution (SR)}
\label{appendix_sr}

\begin{figure}[t]
    \centering
    \includegraphics[width=0.9\linewidth]{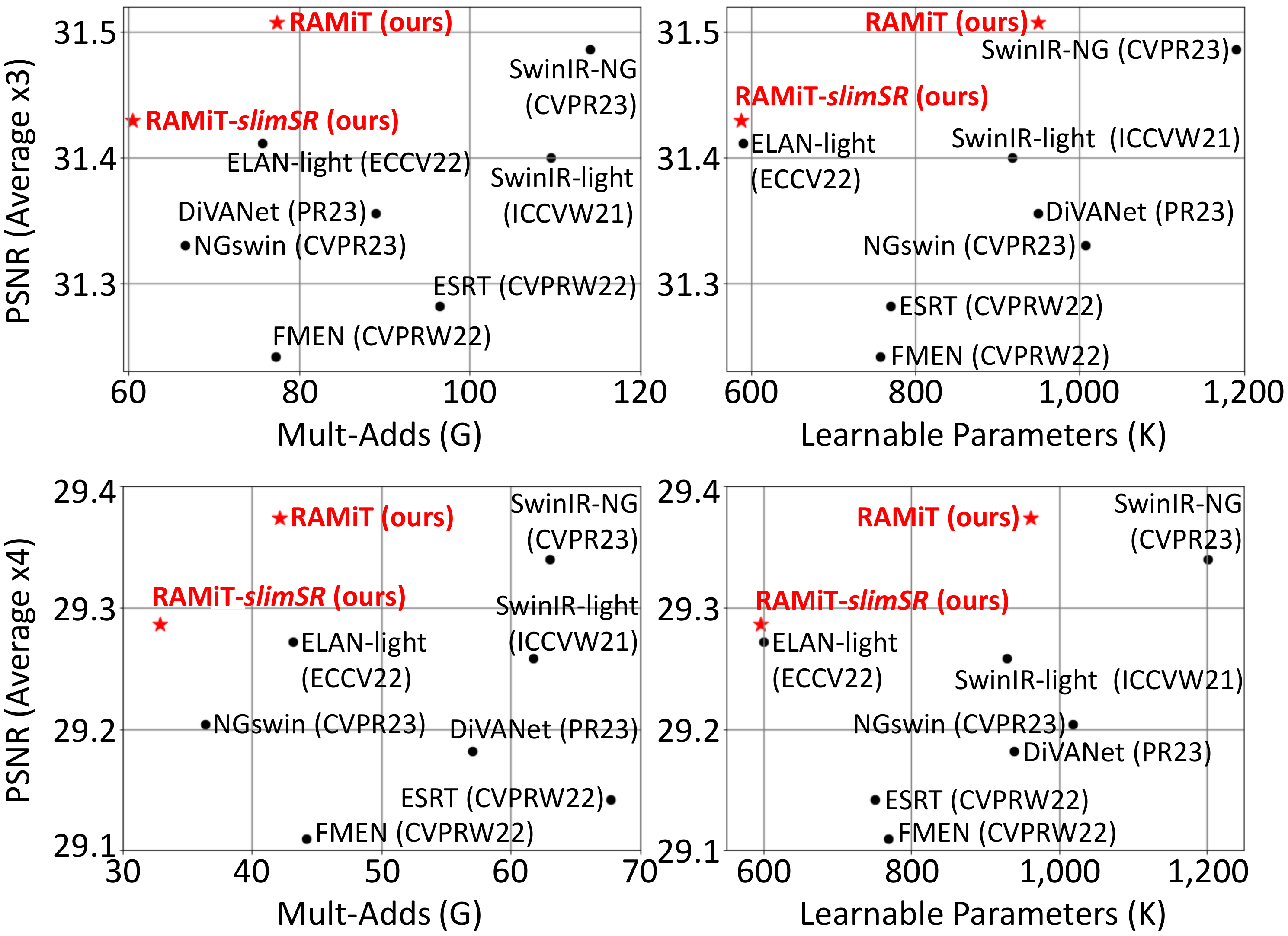} \\
    \caption{Trade-off between efficiency and performance on super-resolution. \textbf{(Top)} $\times3$. \textbf{(Bottom)} $\times4$.}
    \label{appendix_fig_srtradeoff}
\end{figure}

Fig.~\ref{appendix_fig_srtradeoff} illustrate trade-offs between efficiency (Mult-Adds, \#Params) and performance (average PSNR) on SR tasks, including our RAMiT-\textit{slimSR} (Tab.~\ref{appendix_tab_slimsr}) and RAMiT.
Our methods deliver the best trade-off among the comparative models.

\textbf{Smaller Size.}
Tab.~\ref{tab_sr} of the main text appears to have an unfair aspect.
Some networks have fewer parameters than our RAMiT, such as FMEN (CVPRW22)~\cite{du2022fast}, ESRT (CVPRW22)~\cite{Lu_2022_CVPR}, ELAN-light (ECCV22)~\cite{zhang2022efficient}, and DiVANet (PR23)~\cite{behjati2023single}.
Although they require more Mult-Adds than RAMiT, it can be questioned whether our improvement is attributed to the proposed design or the result of having simply more parameters.
We address this issue in Tab.~\ref{appendix_tab_slimsr}.
The channel (network dimension) and depths (D-RAMiT blocks in stage 1 to 4) of RAMiT were scaled from $64$ and $[6,4,4,6]$ to $48$ and $[8,2,2,8]$, respectively.
In the bottleneck and H-RAMi, we also changed the group size and expansion ratio of MobiVari from $4$ and $1.2$ to $1$ and $2.0$, respectively.
The group size and expansion ratio of the other MobiVari layers were retained as the default settings.
Consequently, we got a compact network denoted as RAMiT-\textit{slimSR}, which is composed of the fewest learnable parameters and Mult-Adds among the comparative methods.
Note that RAMiT-\textit{slimSR} consumes fewer computations than NGswin (CVPR23)~\cite{choi2023n}, which required the fewest Mult-Adds in Tab.~\ref{tab_sr}.
RAMiT-\textit{slimSR} still outperformed others, showing that our advancements on super-resolution were attributed to the effectiveness and efficiency of the novel approaches.

\textbf{Training Dataset.}
As shown in Tab.~\ref{appendix_tab_df2k}, we found room for improvement of RAMiT with more training data.
In addition to 800 images of DIV2K~\cite{agustsson2017ntire} used by RAMiT for super-resolution in Tab.~\ref{tab_sr} of the main text, many recent studies utilized 2,650 Flickr2K~\cite{timofte2017ntire} dataset as well to reinforce their SR networks~\cite{liang2021swinir,zhang2021learning,du2022fast,zheng2022cross,zhang2022efficient}.
Following them, we additionally trained our models on DF2K (DIV2K + Flickr2K) for the enhanced performances.
The impacts on all upscaling tasks were observed.

\textbf{Randomness.}
To further prove that the improvements are attributed to not randomness (weight initialization, randomly cropped patches, random data augmentation, etc.) but our approach, we have conducted extra SR experiments as shown in Tab.~\ref{appendix_tab_ablation_random}. RAMiT trained with different random seeds ($\alpha, \beta, \gamma, \delta$) still outperforms SwinIR-NG. The seed $\alpha$ indicates our default.

\begin{table}[t]
    \centering
    \begin{subtable}[h]{\linewidth}
        \caption{Comparison for RAMiT-\textit{slimSR}. The best, second best, and third best results are in \textcolor{red}{red}, \textcolor{orange}{orange}, and \textcolor{blue}{blue}. PSNR and SSIM scores average the results on the five benchmark test datasets.}
        \label{appendix_tab_slimsr}
        \centering
        \resizebox{\linewidth}{!}
        {
        \begin{tabular}{|c|c|c|c|c|c|c||c|c|}
            \hline
            \multirow{2}{*}{} & \multirow{2}{*}{Scale} & \multicolumn{7}{c|}{Method} \\
            \cline{3-9}
            & & FMEN~\cite{du2022fast} & ESRT~\cite{Lu_2022_CVPR} & ELAN-light~\cite{zhang2022efficient} & DiVANet~\cite{behjati2023single} & NGswin~\cite{choi2023n} & \textbf{RAMiT-\textit{slimSR}} & \textbf{RAMiT} \\
            \hline
            Mult-Adds / \#Params & \multirow{2}{*}{$\times2$} & 172.0G / 748K & 191.4G / \textcolor{blue}{677K} & 168.4G / \textcolor{orange}{582K} & 189.0G / 902K & \textcolor{orange}{140.4G} / 998K & \textbf{\textcolor{red}{127.8G}} / \textbf{\textcolor{red}{581K}} & \textbf{\textcolor{blue}{163.4G}} / \textbf{940K} \\
            PSNR / SSIM & & 35.094 / 0.93794 & 35.146 / 0.93754 & \textcolor{orange}{35.258} / \textcolor{orange}{0.93906} & 35.186 / 0.93838 & 35.122 / 0.93836 & \textbf{\textcolor{blue}{35.226}} / \textbf{\textcolor{blue}{0.93880}} & \textbf{\textcolor{red}{35.324}} / \textbf{\textcolor{red}{0.93938}} \\
            \hline
            Mult-Adds / \#Params & \multirow{2}{*}{$\times3$} & 77.2G / \textcolor{blue}{757K} & 96.4G / 770K & \textcolor{blue}{75.7G} / \textcolor{orange}{590K} & 89.0G / 949K & \textcolor{orange}{66.6G} / 1,007K & \textbf{\textcolor{red}{60.4G}} / \textbf{\textcolor{red}{588K}} & \textbf{77.3G} / \textbf{949K} \\
            PSNR / SSIM & & 31.242 / 0.87594 & 31.282 / 0.87586 & \textcolor{blue}{31.412} / \textcolor{blue}{0.87868} & 31.356 / 0.87752 & 31.330 / 0.87778 & \textbf{\textcolor{orange}{31.430}} / \textbf{\textcolor{orange}{0.87872}} & \textbf{\textcolor{red}{31.508}} / \textbf{\textcolor{red}{0.87972}} \\
            \hline
            Mult-Adds / \#Params & \multirow{2}{*}{$\times4$} & 44.2G / 769K & 67.7G / \textcolor{blue}{751K} & 43.2G / \textcolor{orange}{601K} & 57.0G / 939K & \textcolor{orange}{36.4G} / 1,019K & \textbf{\textcolor{red}{32.9G}} / \textbf{\textcolor{red}{597K}} & \textbf{\textcolor{blue}{42.1G}} / \textbf{961K} \\
            PSNR / SSIM & & 29.110 / 0.82376 & 29.142 / 0.82442 & \textcolor{blue}{29.272} / \textcolor{blue}{0.82742} & 29.182 / 0.82570 & 29.204 / 0.82618 & \textbf{\textcolor{orange}{29.286}} / \textbf{\textcolor{orange}{0.82762}} & \textbf{\textcolor{red}{29.374}} / \textbf{\textcolor{red}{0.82940}} \\
            \hline
        \end{tabular}
        }
        \vspace{4pt}
    \end{subtable}
    \begin{subtable}[h]{\linewidth}
        \caption{Training dataset size of RAMiT.}
        \label{appendix_tab_df2k}
        \centering
        \resizebox{\linewidth}{!}
        {
        \begin{tabular}{|c|c|cc|cc|cc|cc|cc||cc|}
            \hline
            Dataset & \multirow{2}{*}{Scale} & \multicolumn{2}{c|}{Set5~\cite{bevilacqua2012low}} & \multicolumn{2}{c|}{Set14~\cite{zeyde2010single}} & \multicolumn{2}{c|}{BSD100~\cite{martin2001database}} & \multicolumn{2}{c|}{Urban100~\cite{huang2015single}} & \multicolumn{2}{c||}{Manga109~\cite{matsui2017sketch}} & \multicolumn{2}{c|}{Average} \\
            \cline{3-14}
            (\#Images) & & PSNR & SSIM & PSNR & SSIM & PSNR & SSIM & PSNR & SSIM & PSNR & SSIM & PSNR & SSIM \\
            \hline
            DIV2K (800) & \multirow{2}{*}{$\times2$} & 38.16 & 0.9612 & \textcolor{red}{34.00} & 0.9213 & 32.33 & 0.9015 & 32.81 & 0.9346 & 39.32 & 0.9783 & 35.324 & 0.93938 \\
            DF2K (3,450) & & \textcolor{red}{38.19} & \textcolor{red}{0.9613} & 33.95 & \textcolor{red}{0.9215} & \textcolor{red}{32.35} & \textcolor{red}{0.9017} & \textcolor{red}{32.90} & \textcolor{red}{0.9352} & \textcolor{red}{39.44} & \textcolor{red}{0.9788} & \textcolor{red}{35.366} & \textcolor{red}{0.93970} \\
            \hline
            DIV2K (800) & \multirow{2}{*}{$\times3$} & 34.63 & 0.9290 & \textcolor{red}{30.60} & 0.8467 & 29.25 & 0.8093 & 28.76 & 0.8646 & 34.30 & 0.9490 & 31.508 & 0.87972 \\
            DF2K (3,450) & & \textcolor{red}{34.69} & \textcolor{red}{0.9295} & \textcolor{red}{30.60} & \textcolor{red}{0.8468} & \textcolor{red}{29.28} & \textcolor{red}{0.8097} & \textcolor{red}{28.80} & \textcolor{red}{0.8656} & \textcolor{red}{34.40} & \textcolor{red}{0.9494} & \textcolor{red}{31.554} & \textcolor{red}{0.88020} \\
            \hline
            DIV2K (800) & \multirow{2}{*}{$\times4$} & 32.56 & 0.8992 & 28.83 & 0.7873 & 27.71 & 0.7418 & 26.60 & 0.8017 & 31.17 & 0.9170 & 29.374 & 0.82940 \\
            DF2K (3,450) & & \textcolor{red}{32.58} & \textcolor{red}{0.8995} & \textcolor{red}{28.87} & \textcolor{red}{0.7876} & \textcolor{red}{27.73} & \textcolor{red}{0.7419} & \textcolor{red}{26.65} & \textcolor{red}{0.8036} & \textcolor{red}{31.25} & \textcolor{red}{0.9174} & \textcolor{red}{29.416} & \textcolor{red}{0.83000} \\
            \hline
        \end{tabular}
        }
    \end{subtable}
    \caption{Ablation study on model size and training dataset for super-resolution.}
    \label{appendix_tab_sr}
\end{table}

\begin{table}[t]
    \centering
    \resizebox{\linewidth}{!}
    {
    \begin{tabular}{|l|c|c|c|c|c|}
        \hline
        Method (seed) & Set5~\cite{bevilacqua2012low} & Set14~\cite{zeyde2010single} & BSD100~\cite{martin2001database} & Urban100~\cite{huang2015single} & Manga109~\cite{matsui2017sketch} \\
        \hline
        SwinIR-NG ($\alpha$) & 38.17 / 34.64 / 32.44 & 33.94 / 30.58 / 28.83 & 32.31 / 29.24 / 27.73 & 32.78 / 28.75 / 26.61 & 39.20 / 34.22 / 31.09 \\
        \hline
        RAMiT ($\alpha$) & 38.16 / 34.63 / {\bf 32.56} & {\bf 34.00} / {\bf 30.60} / {\bf 28.83} & {\bf 32.33} / {\bf 29.25} / 27.71 & {\bf 32.81} / {\bf 28.76} / 26.60 & {\bf 39.32} / {\bf 34.30} / {\bf 31.17} \\
        RAMiT ($\beta$) & {\bf 38.18} / {\bf 34.65} / {\bf 32.54} & {\bf 34.02} / {\bf 30.62} / {\bf 28.86} & {\bf 32.33} / {\bf 29.25} / 27.72 & {\bf 32.81} / {\bf 28.75} / {\bf 26.62} & {\bf 39.28} / {\bf 34.28} / {\bf 31.15} \\
        RAMiT ($\gamma$) & {\bf 38.18} / {\bf 34.64} / {\bf 32.48} & {\bf 34.00} / {\bf 30.60} / 28.80 & {\bf 32.33} / {\bf 29.25} / 27.71 & {\bf 32.83} / {\bf 28.75} / 26.58 & {\bf 39.28} / {\bf 34.29} / {\bf 31.11} \\
        RAMiT ($\delta$) & {\bf 38.18} / {\bf 34.64} / {\bf 32.54} & {\bf 34.02} / {\bf 30.59} / {\bf 28.83} & {\bf 32.32} / {\bf 29.25} / 27.71 & {\bf 32.79} / 28.70 / 26.57 & {\bf 39.27} / {\bf 34.31} / {\bf 31.12} \\
        \hline
    \end{tabular}
    }
    \caption{Ablation on randomness. PSNR on x2 / x3 / x4. The {\bf bold} face indicates better performance over SwinIR-NG~\cite{choi2023n}.}
    \label{appendix_tab_ablation_random}
\end{table}

\textbf{Comparison with Large Model.}
One might question the efficiency of our proposed lightweight method compared to its larger counterpart.
To address this concern, we present Tab.~\ref{appendix_tab_comp_large} where the SwinIR~\cite{liang2021swinir} large model outperforms ours with 12.5 times more parameters than our RAMiT.
However, there is a significant difference in the number of frames per second that SwinIR and RAMiT can process.
Our lightweight method demonstrates superior processing speed in image restoration tasks on both outdated (TITAN Xp) and recent (RTX 4090) GPU devices, surpassing the SwinIR large model.
The result apparently demonstrates that the recent state-of-the-art image restoration models cannot be applied to real-world application despite their enhanced performance.
In contrast, our lightweight approach is specially designed to resolve this efficiency-effectiveness trade-off issue, offering a viable solution for practical implementation.

\begin{table}[t]
    \centering
    \resizebox{\linewidth}{!}
    {
    \begin{tabular}{|l|c|c|c|}
        \hline
        Method & \#Params & Urban100 (PSNR / FPS) & Manga109 (PSNR / FPS) \\
        \hline
        SwinIR~\cite{liang2021swinir} & 11,753K & 33.40 / 0.34, 0.94 & 39.60 / 0.26, 0.71 \\
        RAMiT (ours) & 940K & 32.81 / 1.38, 9.38 & 39.32 / 1.10, 7.38 \\
        \hline
    \end{tabular}
    }
    \caption{Comparison between large and lightweight models. ``FPS'' indicates frames per second processed by each method, which means the higher FPS, the faster, \emph{i.e.}, the better. The former of FPS is measured on an NVIDIA TITAN Xp, while the latter on an NVIDIA GeForce RTX 4090.}
    \label{appendix_tab_comp_large}
\end{table}
\subsection{Low-Light Enhancement (LLE)}
\label{appendix_lle}
\begin{figure*}[t]
    \centering
    \includegraphics[width=0.95\linewidth]{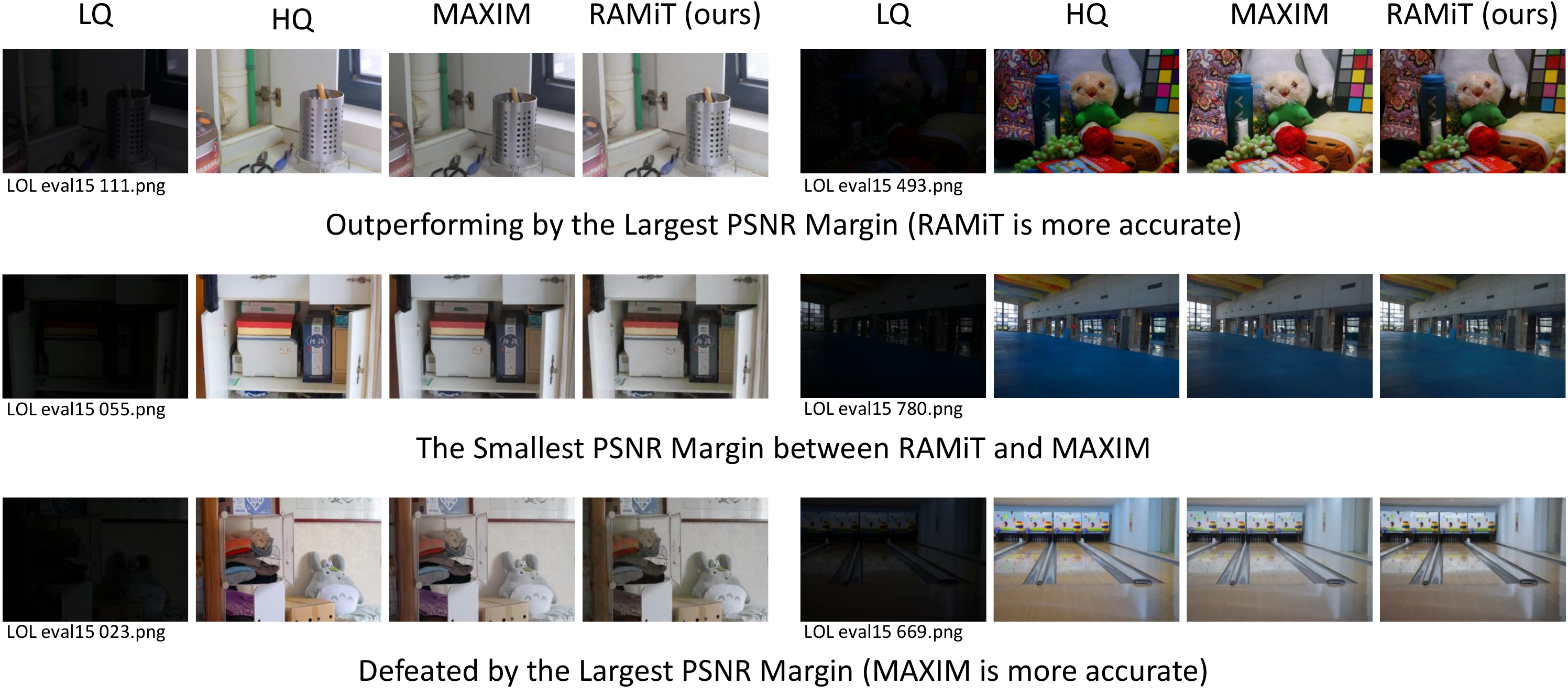} \\
    \caption{Visual comparisons of MAXIM~\cite{tu2022maxim} and RAMiT. Despite even fewer parameters, RAMiT can restore the extremely dark images with better or matched accuracy compared to MAXIM. In the bottom row, the cases in which RAMiT is highly defeated by MAXIM are provided as well.}
    \label{appendix_fig_maxim}
    \vspace{-5pt}
\end{figure*}

Tab.~\ref{appendix_tab_maxim} compares MAXIM (CVPR22)~\cite{tu2022maxim} and RAMiT to present the effectiveness and efficiency of our model for the LLE task.
MAXIM has shown outstanding results on the general image restoration tasks with a large model size.
Surprisingly, our RAMiT outperformed MAXIM in terms of average PSNR scores on the 15 images LOL evaluation dataset~\cite{Chen2018Retinex}.
Notably, we achieved this impressive result using only 6.63\% parameters of MAXIM.
Additionally, RAMiT showed lower variance for the evaluated images than MAXIM, indicating more stable restoration of dark images into brighter ones.
The visual results are in Fig.~\ref{appendix_fig_maxim}.

\begin{table*}[t]
    \centering
    \resizebox{\linewidth}{!}
    {
    \begin{tabular}{|lr|ccccccccccccccc|cc|}
        \hline
        Model & \#Params & 001 & 022 & 023 & 055 & 079 & 111 & 146 & 179 & 493 & 547 & 665 & 669 & 748 & 778 & 780 & Mean & Std. \\
        \hline
        MAXIM~\cite{tu2022maxim} & 14,100K & \textcolor{red}{20.98} & \textcolor{red}{28.68} & \textcolor{red}{24.89} & \textcolor{red}{18.83} & 27.16 & 17.82 & 23.30 & 19.65 & 13.79 & 15.66 & \textcolor{red}{28.34} & \textcolor{red}{28.63} & \textcolor{red}{29.96} & \textcolor{red}{25.02} & \textcolor{red}{28.51} & 23.41 & 5.11 \\
        \textbf{RAMiT (ours)} & \textcolor{red}{935K} & 20.50 & 26.20 & 19.34 & 18.74 & \textcolor{red}{28.18} & \textcolor{red}{31.12} & \textcolor{red}{25.74} & \textcolor{red}{23.61} & \textcolor{red}{20.39} & \textcolor{red}{18.32} & 26.67 & 25.17 & 28.07 & 21.76 & 28.32 & \textcolor{red}{24.14} & \textcolor{red}{3.93} \\
        \hline
    \end{tabular}
    }
    \caption{Comparison of MAXIM~\cite{tu2022maxim} and RAMiT on low-light enhancement. The PSNR (dB) scores on 15 LOL~\cite{Chen2018Retinex} evaluation images are reported. The numbers in the first row indicate the testing file (\texttt{.png}) names. Std.: standard-deviation.}
    \label{appendix_tab_maxim}
\end{table*}

\begin{table}[t]
    \centering
    \begin{subtable}[h]{\linewidth}
        \caption{Comparison for LLE.}
        \label{appendix_tab_slimlle}
        \centering
        \resizebox{\linewidth}{!}
        {
        \begin{tabular}{|r|c|c|c|c|c|}
            \hline
            \multirow{2}{*}{Method} & \multirow{2}{*}{\#Params} & \multicolumn{2}{|c|}{LOL~\cite{Chen2018Retinex}} & \multicolumn{2}{|c|}{VE-LOL-cap~\cite{liu2021benchmarking}} \\
            \cline{3-6}
            & & PSNR & SSIM & PSNR & SSIM \\
            \hline
            URetinex-Net~\cite{wu2022uretinex} & 361K & 21.33 & 0.8348 & 21.22 & 0.8593 \\
            \textbf{RAMiT-\textit{slimLLE} (ours)} & \textbf{\textcolor{red}{358K}} & \textbf{\textcolor{red}{23.77}} & \textbf{\textcolor{red}{0.8379}} & \textbf{\textcolor{red}{28.38}} & \textbf{\textcolor{red}{0.8835}} \\
            \hline
        \end{tabular}
        }
    \end{subtable}
    \begin{subtable}[h]{\linewidth}
        \vspace{3pt}
        \caption{Comparison for DR.}
        \label{appendix_tab_derain}
        \centering
        \resizebox{\linewidth}{!}
        {
        \begin{tabular}{|r|r|c|c|c|c|}
            \hline
            \multirow{2}{*}{Method} & \multirow{2}{*}{\#Params} & \multicolumn{2}{|c|}{Test100~\cite{zhang2019image}} & \multicolumn{2}{|c|}{Rain100H~\cite{yang2017deep}} \\
            \cline{3-6}
            & & PSNR & SSIM & PSNR & SSIM \\
            \hline
            MPRNet~\cite{Zamir2021MPRNet} & 3,637K & 30.27 & 0.8970 & \textcolor{red}{30.41} & \textcolor{red}{0.8990} \\
            \textbf{RAMiT (ours)} & \textbf{\textcolor{red}{935K}} & \textbf{\textcolor{red}{30.44}} & \textbf{\textcolor{red}{0.9012}} & \textbf{29.69} & \textbf{0.8775} \\
            \hline
        \end{tabular}
        }
    \end{subtable}
    \caption{Further comparisons for LLE and DR. \textbf{(a)} RAMiT-\textit{slimLLE} is still better than URetinex-Net. \textbf{(b)} We outperform MPRNet on a benchmark dataset despite much fewer parameters.}
    \label{appendix_tab_uretinex_derain}
\end{table}

Secondarily, we reported a fair comparison with URetinex-Net (CVPR22)~\cite{jiang2021enlightengan} in Tab.~\ref{appendix_tab_slimlle}.
This method requires only 38.6\% parameters of RAMiT, which can provoke a concern of unfairness.
To handle this issue, the channel (network dimension) and the depths (D-RAMiT blocks in stage 1 to 4) of RAMiT were reduced from $64$ to $48$ and from $[6,4,4,6]$ to $[4,2,2,4]$, respectively.
In the bottleneck and H-RAMi, the group size of MobiVari is changed from $4$ to $3$.
As a result, we obtained a downsized model composed of fewer parameters than URetinex-Net, and called it RAMiT-\textit{slimLLE}.
RAMiT-\textit{slimLLE} still outperformed URetinex-Net by PSNR margins of up to 7.16dB, which emphasizes our effectiveness and efficiency.
\subsection{Deraining (DR)}
\label{appendix_dr}
Tab.~\ref{appendix_tab_derain} shows our efficiency for deraining task.
MPRNet (CVPR21)~\cite{Zamir2021MPRNet} made advancements on multiple image restoration tasks a few years ago.
However, RAMiT can outperform it with 25.7\% parameters of MPRNet on a deraining benchmark dataset, such as Test100~\cite{zhang2019image}.
\section{Experimental Details}
\label{appendix_setting}
\textbf{In Common.}
As explained in Sec.~\ref{setting}, we optimized $L_1$ pixel-loss between $I_{RC}$ and $I_{HQ}$ with the Adam optimizer~\cite{kingma2014adam} ($\beta_1=0.9, \beta_2=0.99,\epsilon=10^{-8}$), where $I_{RC}$ is a reconstructed image and $I_{HQ}$ is a high-quality ground-truth image.
Learning rate was initialized as $0.0004 \times 64 / \texttt{batch\_size}$.
The data augmentation method for LQ and HQ pairs was already specified in the main contents.
Before we fed the LQ input images to the network, each input was normalized using \texttt{mean} and \texttt{std} pre-calculated from the LQ training datasets corresponding to each task.
Note that since we used the random (blind) noise levels ($\sigma$) for training our denoising networks, we used \texttt{mean} and \texttt{std} of HQ training datasets for color and grayscale denoising.
When computing the training loss, the normalized $I_{RC}$ was de-normalized (opposite process of normalization).
For evaluation, an input image $I_{LQ}$ was upsized by symmetric padding to fit the size to a multiplier ($=32=8\times2^2$) of the local-window $M(=8)$ and downsizing number ($=2^2$) for the hierarchical stages.
We implemented all processes using PyTorch and two NVIDIA GeForce RTX 4090 GPUs.
The implementation details of RAMiT are in Tab.~\ref{appendix_tab_implement}.

\begin{table}[ht]
    \centering
    \resizebox{\linewidth}{!}
    {
    \begin{tabular}{|c|r|l|}
        \hline
        & dim ($C$) & $64$ \\
        \multirow{2}{*}{Overall} & depths & $[6,4,4,6]$ \\
        \multirow{2}{*}{Architecture} & num heads & $[4,4,4,4]$ \\
        & chsa head ratio ($L_{ch}/L$) & $25\%$ \\
        & window size ($M$) & $8$ \\
        \hline
        Feed-Forward & hidden ratio & $2.0$ \\
        Network (FFN) & activation & GELU~\cite{hendrycks2016gaussian} \\
        \cline{1-3}
        \multirow{3}{*}{MobiVari} & exp factor & $1.2$ \\
        & expand groups & $4$ \\
        & activation & LeakyReLU~\cite{maas2013rectifier} \\
        \hline
        \multirow{3}{*}{Dropout} & attention map & $0.0$ \\
        & attention project & $0.0$ \\
        & drop path  & $0.0$ \\
        \hline
        \multirow{5}{*}{Others} & optimizer & Adam~\cite{kingma2014adam} ($\beta_1=0.9, \beta_2=0.99,\epsilon=10^{-8}$) \\
        & initialized learning rate & $0.0004 \times 64 / \texttt{batch\_size}$ \\
        & learning rate decay & half (see paragraphs below) \\
        & batch size & see paragraphs below \\
        & epoch / total datapoints & 500 / 32M (SR), 400 / 10M (Others) \\
        \hline
    \end{tabular}
    }
    \caption{Implementation details of RAMiT. ``depths'' and ``num heads'' count the number of D-RAMiT blocks ($[\mathbb{K}_1,\mathbb{K}_2,\mathbb{K}_3,\mathbb{K}_4]$) and multi-heads ($L$) in stage $1,2,3,4$. Correspondingly, the setting of ``chsa head ratio =$25\%$'' indicates that $(L_{sp}, L_{ch})$ is placed as $[(3,1), (3,1), (3,1), (3,1)]$ in each stage.}
    \label{appendix_tab_implement}
\end{table}

\textbf{Super-Resolution.}
We trained RAMiT for $\times2$ task from scratch, of which the training epochs were set to $500$.
For $\times3,\times4$ tasks, we followed a warm-start strategy~\cite{lin2022revisiting}, where we fine-tuned the final reconstruction module for $50$ epochs (warm-start phase) before fine-tuning whole network parameters (whole-finetuning phase) lasting for $250$ epochs.
In warm-start phase, the network parameters pre-trained on $\times2$ task were loaded to initialize $\times3$, $\times4$ networks, except for the reconstruction module.
Learning rate was decayed by half at $\{200,300,400,425,450,475\}$ and $\{50,100,150,175,200,225\}$ epochs for training from scratch ($\times2$) and whole-finetuning phase ($\times3,\times4$), respectively.
Learning rate of warm-start phase remained as a constant (\emph{i.e.}, $0.0004\times64/\texttt{batch\_size}$).
We also linearly increased learning rate from $0$ to $0.0004\times64/\texttt{batch\_size}$ during the first $20$ epochs of the training from scratch and whole-finetuning phase (warmup epoch~\cite{goyal2017accurate}).
Each training image was cropped into a patch size of $64\times64$ with $64$ batch size regardless of training from scratch or warm-start strategy.
To consistently manage the datapoints per epoch, we repeated each datapoint $80$ and $18.551$ times for DIV2K and DF2K datasets, which made the number of training images used for an epoch equal to $64,000$.

\textbf{Others.}
For color and grayscale denoising, low-light enhancement, and deraining, we adapted the progressive learning~\cite{Zamir2021Restormer}, where the patch size was initially set to $64\times64$, and then progressively increased to $96\times96$ and $128\times128$ after $\{100,200\}$ epochs, respectively.
The corresponding batch size was $\{64,32,16\}$.
We decrease learning rate by half at $\{200,300,350,375\}$ epochs.
Warmup epoch was the same as super-resolution.
The training process lasted for $400$ epochs.
Similar to super-resolution, we repeated each datapoint $3.0$, $14.006$, and $1.8234$ times for denoising, low-light enhancement, and deraining, respectively (about $25,000$ datapoints were used per epoch).
While we obtained the synthetic or real-captured low and high quality image pairs of low-light enhancement and deraining from public sources\footnote{LOL and VE-LOL datasets can be found in this \href{https://daooshee.github.io/BMVC2018website}{\texttt{website1}} and this \href{https://flyywh.github.io/IJCV2021LowLight_VELOL/}{\texttt{website2}}. Deraining Testsets and Rain13K can be publicly downloaded in this \href{https://drive.google.com/drive/folders/1PDWggNh8ylevFmrjo-JEvlmqsDlWWvZs}{\texttt{google-drive1}} and this \href{https://drive.google.com/drive/folders/1Hnnlc5kI0v9_BtfMytC2LR5VpLAFZtVe}{\texttt{google-drive2}}.}, the Additive White Gaussian Noise (\texttt{AWGN}) for low quality noisy input images of denoising tasks was generated by the following PyTorch-like code:

\begin{center}
\begin{tabular}{l}
     \texttt{AWGN = torch.randn(*img\_hq.shape)*$\sigma$/255} \\
     \texttt{img\_lq = img\_hq + AWGN}, \\
\end{tabular}
\end{center}

where the random seed was set to $0$ for the evaluation process (in training, seed is not given to implement blind denoising); \texttt{img\_lq} and \texttt{img\_hq} indicate low and high quality images; $\sigma$ is noise level set to one among $[15,25,50]$ for testing or sampled uniformly between $0\sim50$ for training.
\section{More Visual Comparisons}
\label{appendix_viscomp}
\begin{figure*}[t]
    \centering
    \includegraphics[width=\linewidth]{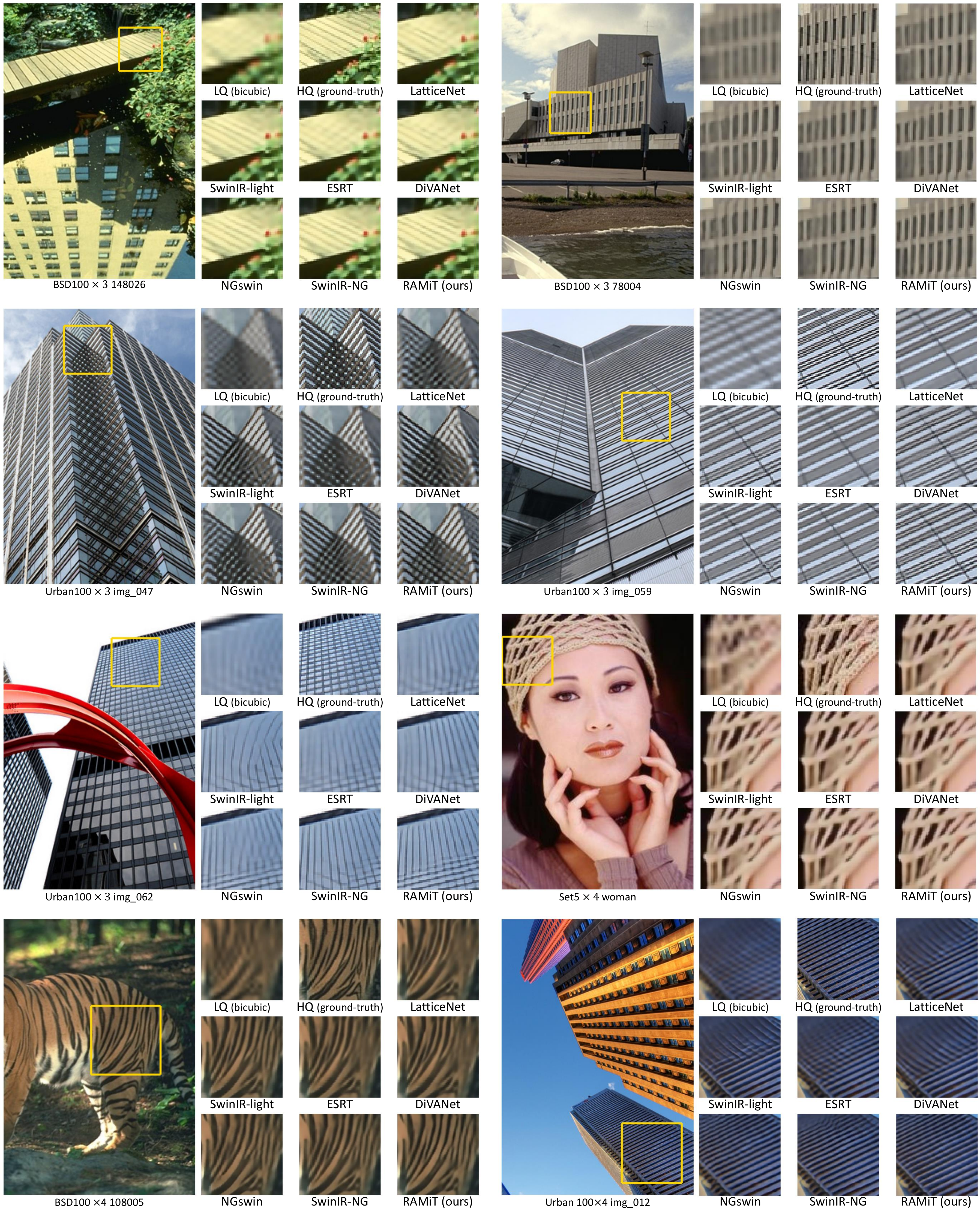} \\
    \caption{Visual comparisons of super-resolution. LQ: Low-Quality input. HQ: High-Quality target.}
    \label{appendix_fig_viscomp_sr1}
\end{figure*}

\begin{figure*}[t]
    \centering
    \includegraphics[width=\linewidth]{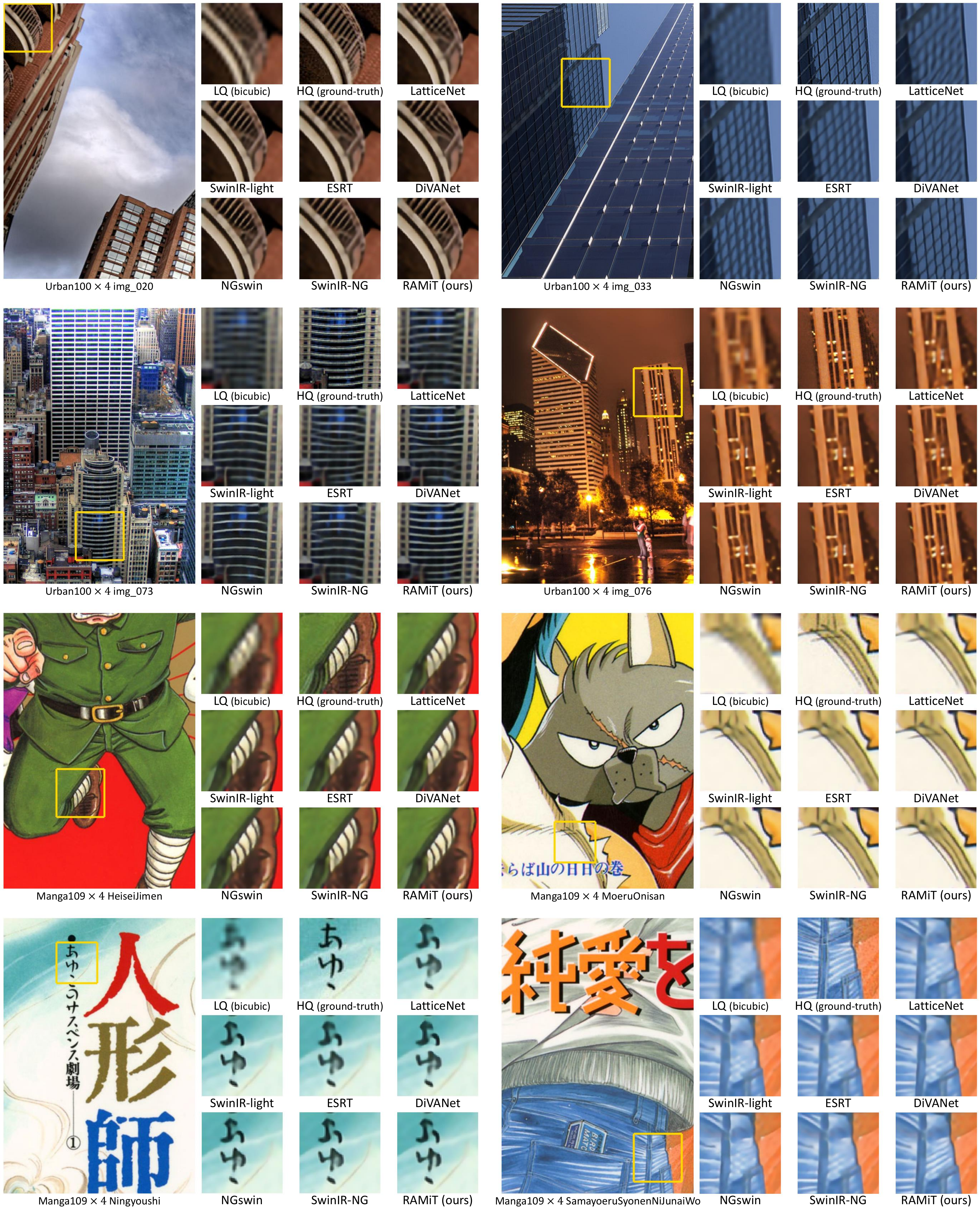} \\
    \caption{Visual comparisons of super-resolution. LQ: Low-Quality input. HQ: High-Quality target.}
    \label{appendix_fig_viscomp_sr2}
\end{figure*}

\begin{figure*}[t]
    \centering
    \includegraphics[width=\linewidth]{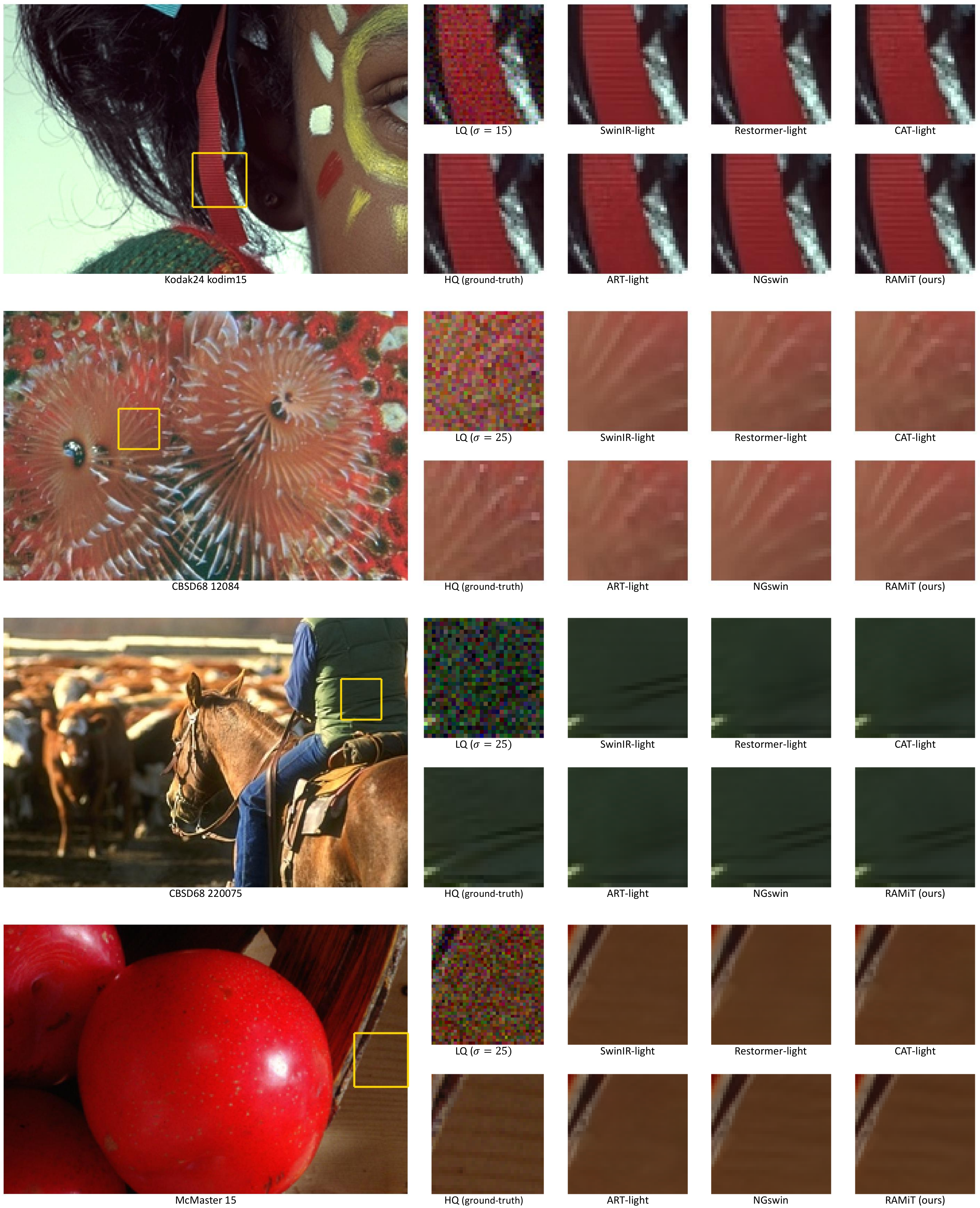} \\
    \caption{Visual comparisons of denoising. LQ: Low-Quality input. HQ: High-Quality target.}
    \label{appendix_fig_viscomp_dn1}
\end{figure*}

\begin{figure*}[t]
    \centering
    \includegraphics[width=\linewidth]{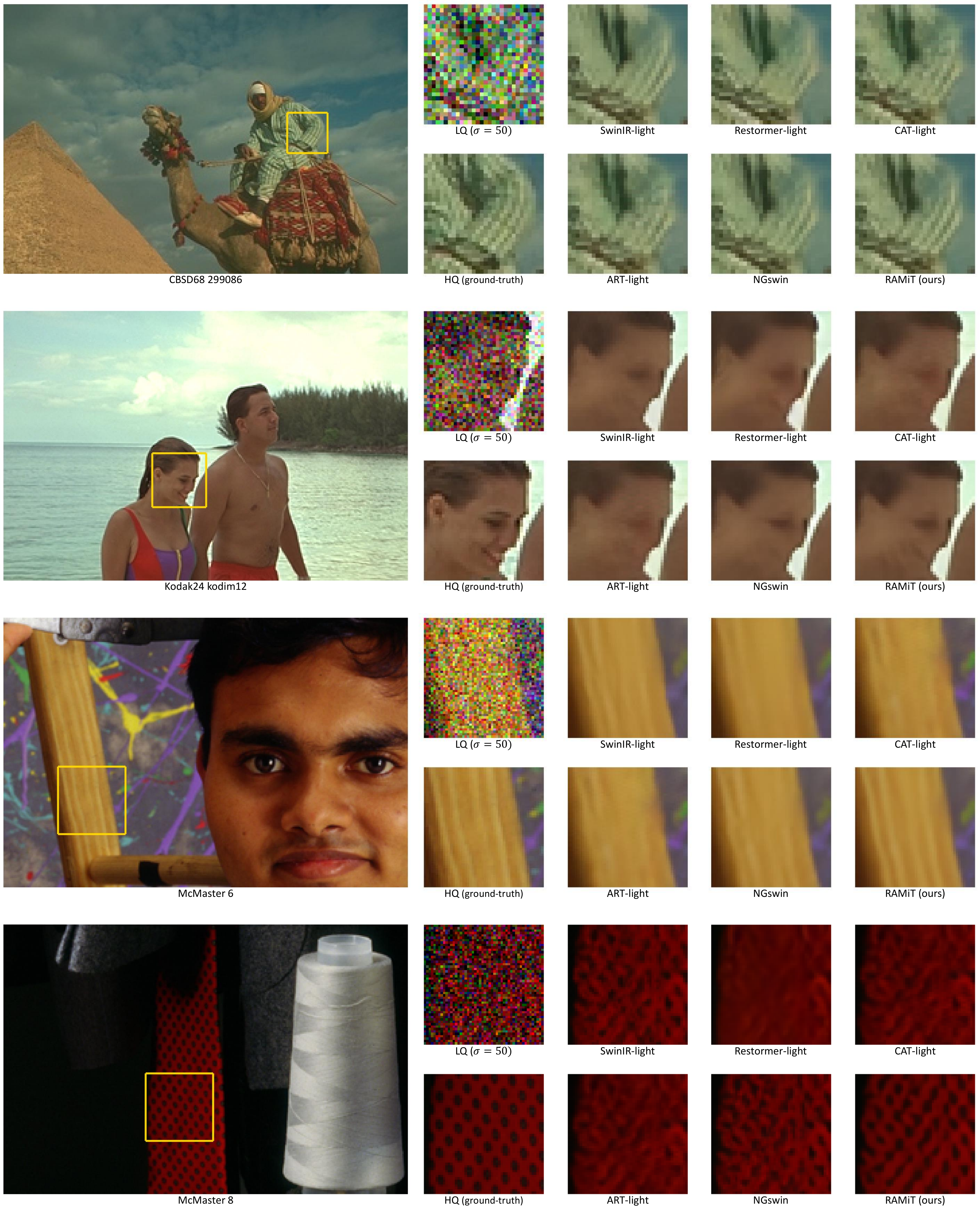} \\
    \caption{Visual comparisons of denoising. LQ: Low-Quality input. HQ: High-Quality target.}
    \label{appendix_fig_viscomp_dn2}
\end{figure*}

\begin{figure*}[t]
    \centering
    \includegraphics[width=\linewidth]{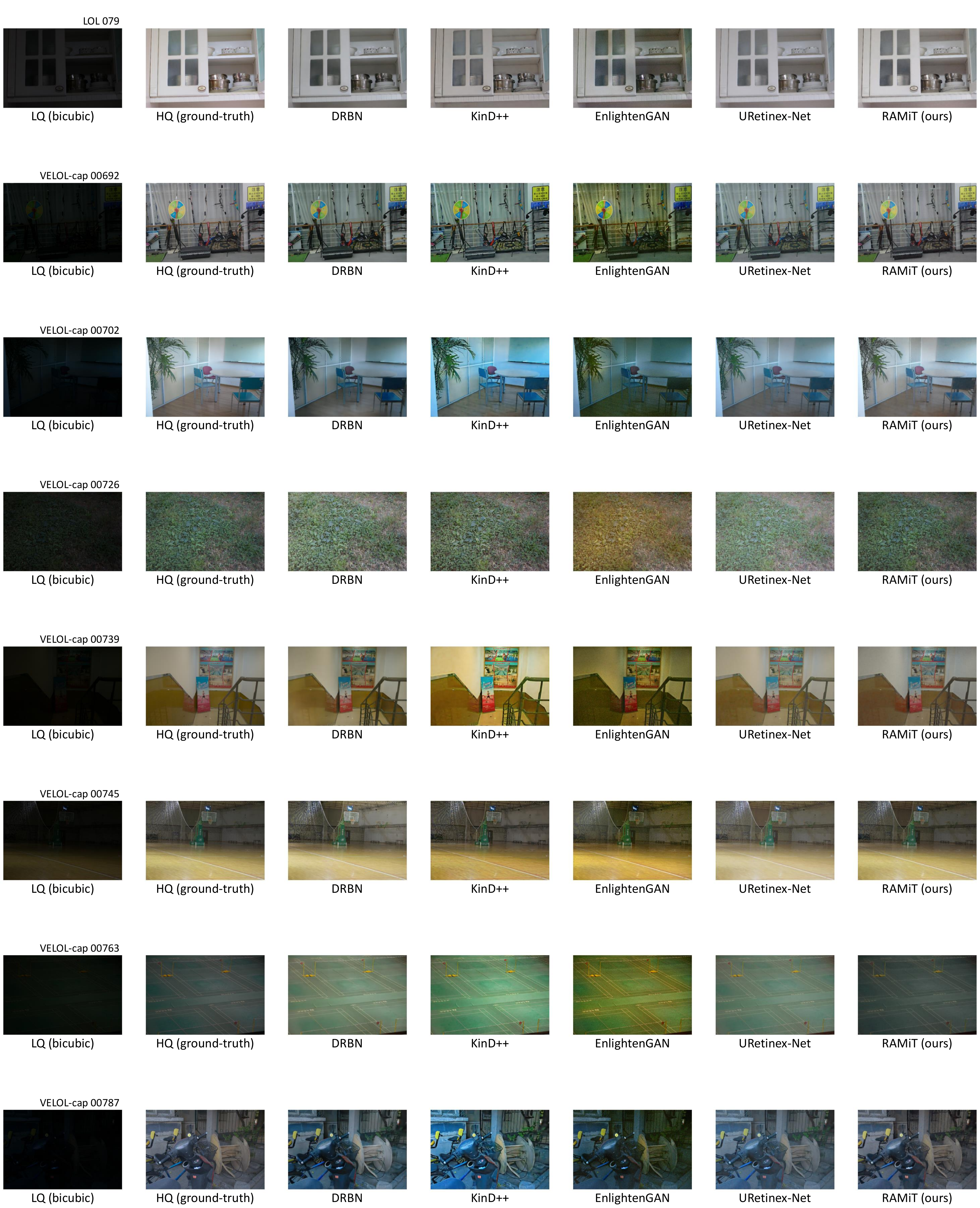} \\
    \caption{Visual comparisons of low-light enhancement. LQ: Low-Quality input. HQ: High-Quality target.}
    \label{appendix_fig_viscomp_lle}
\end{figure*}

\begin{figure*}[t]
    \centering
    \includegraphics[width=0.8\linewidth]{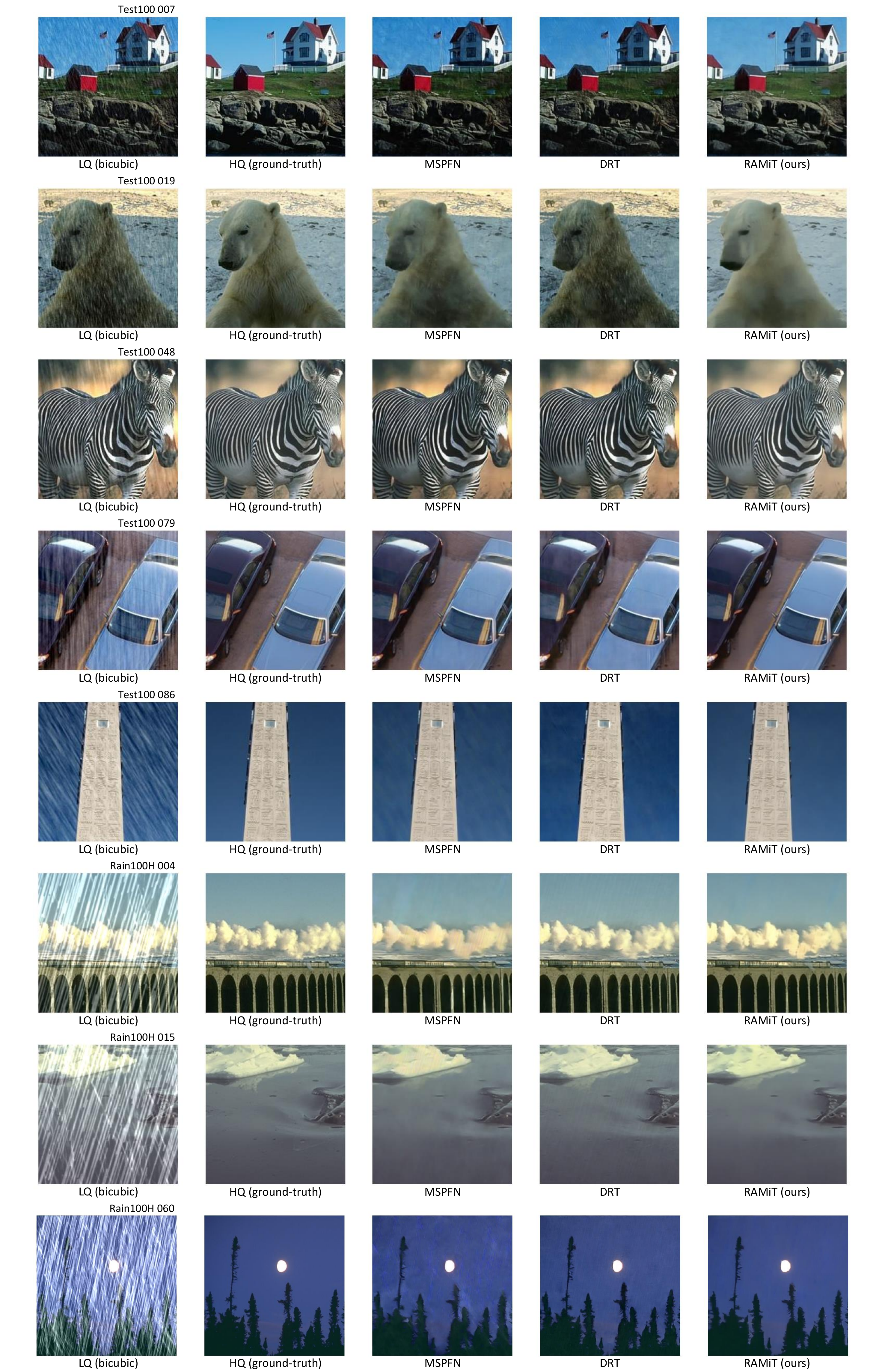} \\
    \caption{Visual comparisons of deraining. LQ: Low-Quality input. HQ: High-Quality target.}
    \label{appendix_fig_viscomp_dr}
\end{figure*}

In the last six pages (P.~\pageref{appendix_fig_viscomp_sr1}--\pageref{appendix_fig_viscomp_dr} after References) of this document, we provide additional visual comparisons of our RAMiT and other networks.
These visual results exhibit the effectiveness of our approach on super-resolution (Figs.~\ref{appendix_fig_viscomp_sr1}~and~\ref{appendix_fig_viscomp_sr2}), denoising (Figs.~\ref{appendix_fig_viscomp_dn1}~and~\ref{appendix_fig_viscomp_dn2}), low-light enhancement (Fig.~\ref{appendix_fig_viscomp_lle}), and deraining (Fig.~\ref{appendix_fig_viscomp_dr}).
{
    \small
    \bibliographystyle{ieeenat_fullname}
    \bibliography{main}
}
\clearpage

\end{document}